%% file: main.tex
\documentclass[10pt,twocolumn,letterpaper]{article}

\usepackage{iccv}
\usepackage{times}
\usepackage{booktabs}
\usepackage{epsfig}
\usepackage{graphicx}
\usepackage{amsmath}
\usepackage{amssymb}
\usepackage{xspace}
\usepackage{xcolor}
\usepackage{tabularx}
\usepackage{multirow}
\usepackage{rotating}
\usepackage{placeins}
\usepackage{paralist}
\usepackage{subcaption}
\usepackage[hang,flushmargin]{footmisc}
\usepackage[accsupp]{axessibility}

\usepackage[pagebackref=true,breaklinks=true,letterpaper=true,colorlinks,bookmarks=false]{hyperref}

\iccvfinalcopy % *** Uncomment this line for the final submission

\input{_notations}

\newcommand{\heading}[1]{
    \vspace{0.05in}
    \noindent{\bf #1}\xspace
}

\renewcommand{\sectionautorefname}{Section}
\let\subsectionautorefname\sectionautorefname
\let\subsubsectionautorefname\sectionautorefname

\begin{document}

%%%%%%%%% TITLE
\title{Video Pose Distillation for Few-Shot, Fine-Grained Sports Action Recognition}

\author{James Hong\textsuperscript{1}
\quad
Matthew Fisher\textsuperscript{2}
\quad
Micha\"el Gharbi\textsuperscript{2}
\quad
Kayvon Fatahalian\textsuperscript{1} \\
\textsuperscript{1}Stanford University
\qquad
\textsuperscript{2}Adobe Research
}

\maketitle
% Remove page # from the first page of camera-ready.
\ificcvfinal\thispagestyle{empty}\fi

\input{sections/abstract}

\input{figures/bad_poses}

%%%%%%%%% BODY TEXT
\input{sections/intro}

\input{sections/related}

\input{sections/method}

\input{sections/result_recognize}

\input{sections/result_other}

\input{sections/discussion_and_conclusion}

\input{sections/ack}

\FloatBarrier
{\small
\bibliographystyle{ieee_fullname}
\bibliography{main}
}

\FloatBarrier
\newpage
\appendix

\renewcommand*{\thesection}{\Alph{section}}
\renewcommand*{\thesubsection}{\Alph{section}\arabic{subsection}}
\renewcommand*{\thesubsubsection}{\Alph{section}\arabic{subsection}.\arabic{subsubsection}}
\let\subsectionautorefname\sectionautorefname
\let\subsubsectionautorefname\sectionautorefname

\input{supplemental/implementation}
\input{supplemental/experiments}
\input{supplemental/dataset}
\input{supplemental/vipestar}

\end{document}

%% file: _notations.tex
\DeclareMathOperator*{\minimize}{minimize}

\newcommand{\notation}[1]{\ensuremath{#1}\xspace}

\newcommand{\OURMETHOD}{{VPD}\xspace}
\newcommand{\VIPESTAR}{{VIPE$^\star$}\xspace}
\newcommand{\fs}{{FSJump6}\xspace}
\newcommand{\tennis}{{Tennis7}\xspace}
\newcommand{\fx}{{FX35}\xspace}

\newcommand{\EmbeddingDim}{\notation{d}}
\newcommand{\PoseDim}{\notation{d}}
\newcommand{\Height}{\notation{h}}
\newcommand{\Width}{\notation{w}}
\newcommand{\ImageDims}{\notation{3\Height\Width}}
\newcommand{\FlowDims}{\notation{2\Height\Width}}
\newcommand{\NumFrames}{\notation{N}}

\newcommand{\Time}{\notation{t}}

\newcommand{\Frame}[1]{\notation{\mathbf{x}_{#1}}}
\newcommand{\Flow}[1]{\notation{\mathbf{\phi}_{#1}}}

\newcommand{\WeakPose}[1]{\notation{\mathbf{p}_{#1}}}

\newcommand{\PoseMotion}[1]{\notation{\Delta\WeakPose{#1}}}

\newcommand{\Student}{\notation{F}}
\newcommand{\PoseDecoder}{\notation{D}}

\newcommand{\textsuperdagger}{\textsuperscript{\textdagger}}

%% file: sections/abstract.tex
\begin{abstract}

Human pose is a useful feature for fine-grained sports action understanding.
However, pose estimators are often unreliable when run on sports video due to domain shift and factors such as motion blur and occlusions.
This leads to poor accuracy when downstream tasks, such as action recognition, depend on pose.
End-to-end learning circumvents pose, but requires more labels to generalize.

We introduce Video Pose Distillation (\OURMETHOD), a weakly-supervised
technique to learn features for new video domains, such as individual sports that challenge pose estimation.
Under \OURMETHOD, a student network learns to extract robust pose
features from RGB frames in the sports video, such that, whenever pose
is considered reliable, the features match the output of a pretrained teacher
pose detector. Our strategy retains the best of both pose and end-to-end worlds, exploiting the
rich visual patterns in raw video frames, while learning features that agree
with the athletes' pose and motion in the target video domain
to avoid over-fitting to patterns unrelated to athletes' motion.

\OURMETHOD features improve performance on few-shot, fine-grained action
recognition, retrieval, and detection tasks in four real-world sports video datasets, without requiring additional ground-truth pose annotations.

\end{abstract}

%% file: figures/bad_poses.tex
\begin{figure}[tbp]
    \centering
    \includegraphics[width=\columnwidth]{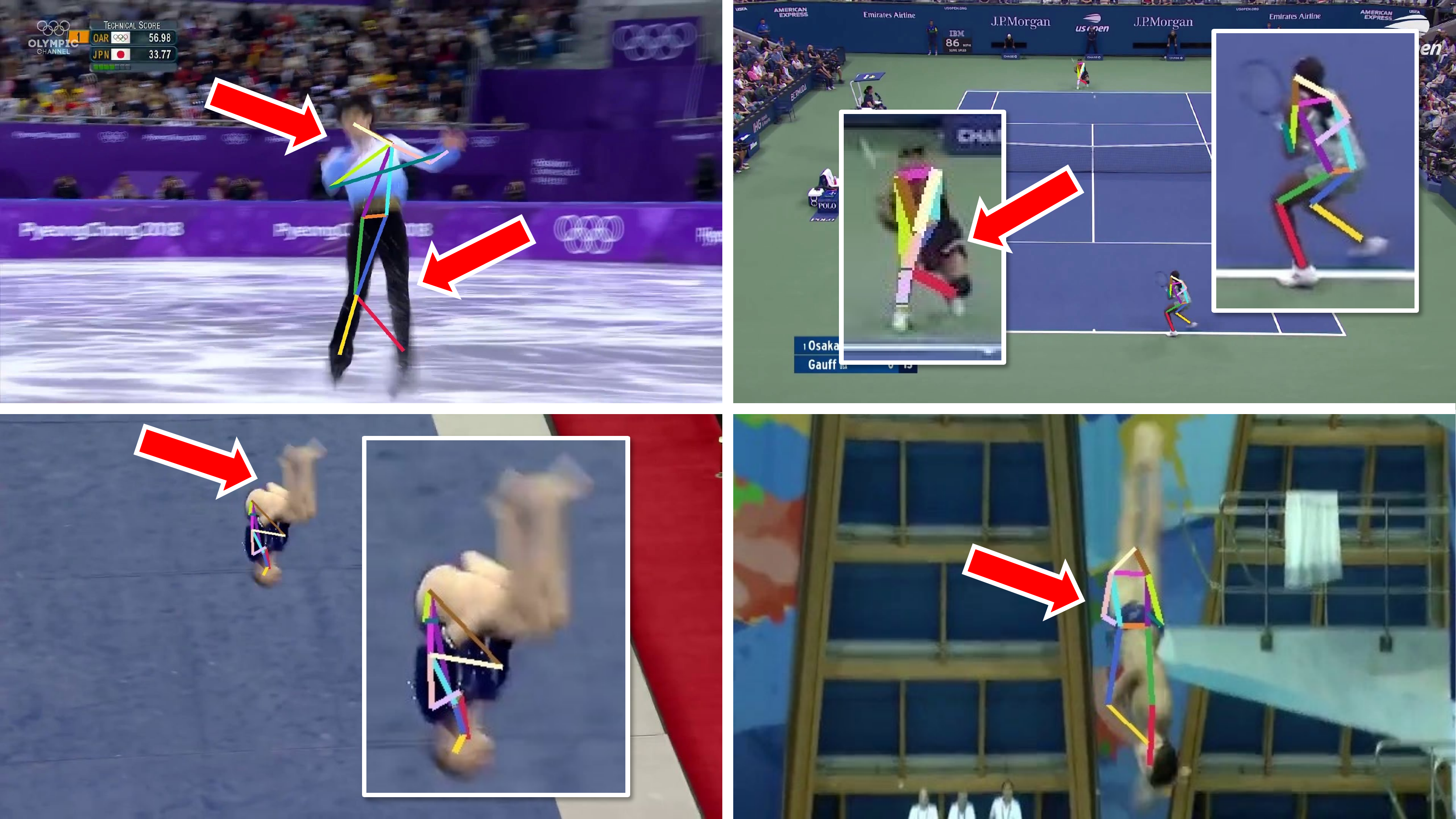}
    \vspace{-0.2in}
    \caption{
        {\bf Limitations of current 2D pose detectors.}
        State-of-the-art pose estimators~\cite{hrnet} produce noisy and
        incorrect results in frames with challenging motions, common in sports
        video.
        Here are examples from figure skating, tennis, gymnastics, and
        diving where pose estimates are incorrect.
    }
    \label{fig:bad_poses}
    \vspace{-0.8em}
\end{figure}

%% file: sections/intro.tex
\section{Introduction}

Analyzing sports video requires robust algorithms to automate
fine-grained action recognition, retrieval, and detection in large-scale video collections.
Human pose is a useful feature when sports are centered around people.

State-of-the-art skeleton-based deep learning techniques for action recognition~\cite{msg3d,stgcn} rely on accurate 2D pose detection to
extract the athletes' motion, but the best pose detectors~\cite{hrnet,detectron2}
routinely fail on fast-paced sports video with complex blur and occlusions, often in frames crucial to the action (\autoref{fig:bad_poses}).
To circumvent these issues, end-to-end learned models operate directly on the video
stream~\cite{i3d,slowfast,tsm,gsm,tsn,trn}.
However, because they consume pixel instead of pose inputs, when trained with few labels, they
tend to latch onto specific visual patterns~\cite{danceinmall,mimetics} rather than the fine-grained motion (e.g., an athlete's clothes or the presence of a ball).
As a result, prior pose and end-to-end methods often generalize poorly on fine-grained tasks in challenging sports video, when labels are scarce.
While collecting large datasets with fine action and pose annotations is possible, doing so for each new sport does not scale.

We propose \textbf{Video Pose Distillation} (\OURMETHOD), a weakly-supervised technique in which a \emph{student} network learns to extract robust pose features from RGB video frames in a new video domain (a single sport).
\OURMETHOD is designed such that, whenever pose
is reliable, the features match the output of a pretrained \emph{teacher}
pose detector.
Our strategy retains the best of both pose and end-to-end worlds.
First, like directly supervised end-to-end methods, our student can exploit the rich
visual patterns present in the raw frames, including but not limited to the
athlete's pose, and continue to operate when pose estimation is unsuccessful.
Second, by constraining our descriptors to agree with the pose estimator
whenever high-confidence pose is available, we avoid the pitfall of overfitting to visual patterns unrelated to the athlete's action.
And third, weak pose supervision allows us to enforce an additional
constraint: we require that the student predicts not only instantaneous pose
but also its temporal derivative.
This encourages our features to pick up on visual similarities over time (e.g., an athlete
progressing from pose to pose).
When we train the student with weak-supervision over a corpus of unlabeled sports video,
the student learns to `fill-in the gaps' left by the noisy pose teacher.
Together, these properties lead to a student network whose features outperform
the teacher's pose output when used in downstream applications.

\OURMETHOD features improve performance on few-shot, fine-grained action
recognition, retrieval, and detection tasks in the target sport domain,
without requiring additional ground-truth action or pose labels.
We demonstrate the benefits of \OURMETHOD on four diverse sports video datasets with {\em fine-grained} action labels: diving~\cite{diving48}, floor exercises~\cite{finegym}, tennis~\cite{vid2player}, and a new dataset for figure skating.
In a few-shot --- limited supervision --- setting, action recognition models trained with distilled \OURMETHOD features
can significantly outperform models trained directly on features from the teacher as well as baselines from prior skeleton-based and end-to-end learning work.
For instance, when restricted to between 8 and 64 training examples per class from diving and floor exercises, the two datasets that are most challenging for pose, \OURMETHOD features improve fine-grained classification accuracy by 6.8 to 22.8\% and by 5.0 to 10.5\%, respectively, over the next best method(s).
Even when labels are plentiful, \OURMETHOD remains competitive, achieving superior accuracy on three of the four test datasets.
To summarize, \OURMETHOD surpasses its teacher in situations where leveraging pose is crucial (e.g., few-shot) and is also competitive when end-to-end methods dominate (e.g., unreliable pose and the high-data / full supervision setting).
Finally, we show applications of \OURMETHOD features to fine-grained action retrieval and few-shot temporal detection tasks.

This paper makes the following contributions:
\begin{enumerate}
\item A weakly-supervised method, \OURMETHOD, to adapt pose features to
    new video domains, which significantly improves performance on downstream tasks
    like action recognition, retrieval, and detection in scenarios where 2D
    pose estimation is unreliable.
\item State-of-the-art accuracy in few-shot, fine-grained action understanding
    tasks using \OURMETHOD features, for a variety of sports. On action recognition,
    \OURMETHOD features perform well with as few as 8 examples per class and remain competitive or state-of-the-art even as the training data is increased.
\item
    A new dataset (figure skating) and extensions to three datasets of real-world sports video,
    to include tracking of the performers, in order to facilitate future research on
    fine-grained sports action understanding.
\end{enumerate}

%% file: sections/related.tex
\section{Related Work}

\noindent\textbf{Pose representations} provide a powerful abstraction for
human action understanding.
Despite significant progress in 2D and 3D pose
estimation~\cite{personlab,videopose3d,hrnet}, downstream algorithms that
depend on pose continue to suffer from unreliable estimates in sports video.
With few labels available, for tasks such as fine-grained action recognition,
models must learn both the actions and to cope with noisy inputs.

VIPE~\cite{prvipe} and CV-MIM~\cite{cvmim} show that learned pose
embeddings, which factor-out camera view and
forgo explicit 3D pose estimation, can be useful; they are
trained on out-of-domain 3D pose data to embed 2D pose inputs and are effective when 2D pose is reliable.
\OURMETHOD extends these works by using distillation to replace the unreliable 2D pose estimation step with a model that embeds directly from pixels to pose-embedding.
\cite{learnhumanmotion,videopose3d,phd} learn human motion from video but produce 3D pose rather than embeddings.

\input{figures/system_diagram}

\heading{Video action recognition} is dominated by end-to-end models~\cite{timesformer,i3d,slowfast,tsm,gsm,r3d,tsn,trn}, which are often evaluated on diverse but coarse-grained classification tasks (e.g., `golf', `tennis', etc.)~\cite{kinetics,hmdb51,minimit,ucf101,pennaction}.
Fine-grained action recognition in sports is a recent development~\cite{diving48,finegym}.
Besides being necessary for sports video analysis, fine-grained classification within a single sport is interesting because it avoids many contextual biases in coarse-grained tasks~\cite{danceinmall,diving48,mimetics}.
\cite{ikea,epickitchens,ssv1,ssv2} are also fine-grained datasets, but differ from body-centric actions in sports.

Pose or skeleton-based methods~\cite{potion,msg3d,stgcn} appear to be a good fit for action recognition in human-centric sports.
They depend on reliable 2D or 3D pose, which exists in datasets captured in controlled settings~\cite{nturgbd120,nturgbd} but not for public sports video, where no ground-truth is available and automatic detectors often perform poorly (e.g.,~\cite{diving48,finegym}).

\OURMETHOD improves upon pose-based and end-to-end methods in human-centric sports datasets, especially when pose is not reliable.
Like VIPE~\cite{prvipe}, \OURMETHOD produces effective pose features, to the extent that comparatively simple downstream models such as nearest neighbor search~\cite{prvipe} or a generic BiGRU~\cite{rnnchapter} network can compete with the state-of-the-art in action recognition --- in both few-shot and high-data regimes.
To show this, we compare against several recent action recognition methods~\cite{msg3d,gsm} in~\autoref{sec:action_recognition}.

\OURMETHOD features can be used for any tasks where pretrained pose features may be helpful, such as action retrieval and temporally fine-grained detection (e.g., identifying tennis racket swings at 200 ms granularity).
The latter is interesting because prior baselines~\cite{activitynet,thumos14} focus on more general categories than human-centric action within a single sport and few papers~\cite{taen,detectionsimilarity} address the few-shot setting.

\heading{Few-shot action recognition} literature follows a number of paradigms, including meta-learning, metric learning, and data-augmentation approaches~\cite{taen,temporalalignment,protogan,generativefewshot}.
These works focus on coarse-grained datasets~\cite{activitynet,kinetics,hmdb51,ucf101}, adopt various protocols that partition the dataset into seen/unseen classes and/or perform a reduced N-way, K-shot classification (e.g., 5-way, 1 or 5 shot).
\OURMETHOD differs in that it is completely agnostic to action labels when training features and does not require a particular architecture for downstream tasks such as action recognition.
In contrast to `few-shot' learning that seeks to generalize to unseen classes, {\em we evaluate on the standard classification task, with all classes known, but restricted to only $k$-examples per class at training time.}
Our evaluation is similar to~\cite{fixmatch,cvmim}, which perform action and image recognition with limited supervision, and, like~\cite{fixmatch,cvmim}, we test at different levels of supervision.

\heading{Self-supervision/distillation.}
\OURMETHOD relies on only machine-generated pose annotations for weak-supervision and distillation.
\OURMETHOD is similar to~\cite{noisystudentimagenet} in that the main goal of distillation is to improve the robustness and accuracy of the student rather than improve model efficiency.
Most self-supervision work focuses on pretraining and joint-training scenarios, where self-supervised losses are secondary to the end-task loss, and subsequent or concurrent fine-tuning is necessary to obtain competitive results~\cite{simclr,coclr,temporaltransform,cubicpuzzles,ms2l}.
By contrast, our \OURMETHOD student is fixed after distillation.

%% file: figures/system_diagram.tex
\begin{figure*}[tbp]
    \centering
    \includegraphics[width=\textwidth]{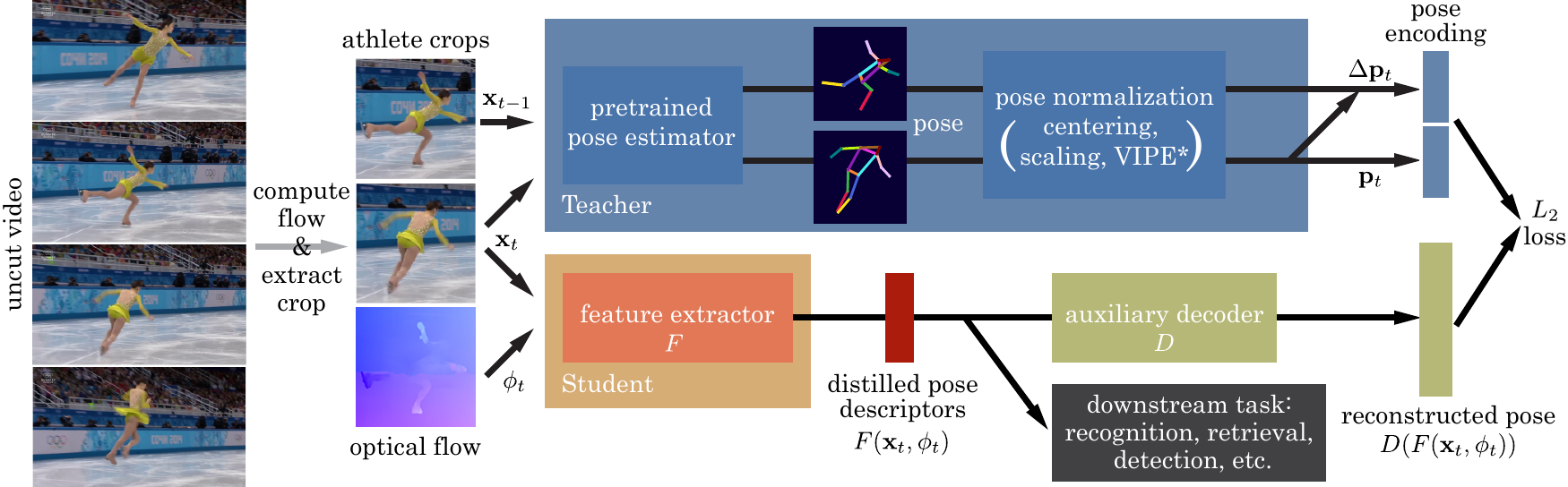}
    \vspace{-0.25in}
    \caption{\label{fig:system_diagram}
        {\bf Method overview.}
        \OURMETHOD has two data pathways: a teacher
        to generate supervision and a student that learns to embed pose and motion in
        the target (sport) domain.
        When training on a frame \Time, the teacher applies an
        off-the-shelf 2D pose estimator, followed by a pose normalization step,
        to obtain weak pose features: \WeakPose{\Time} and \PoseMotion{\Time}.
        The student pathway receives the localized RGB \Frame{\Time} and optical
        flow \Flow{\Time}, and computes a descriptor
        $\Student\left(\Frame{\Time},\Flow{\Time}\right)\in\mathbb{R}^\EmbeddingDim$,
        from which the fully connected network, \PoseDecoder, regresses
        $\left(\WeakPose{\Time}, \PoseMotion{\Time}\right)$.
        After training, only \Student is retained to extract embeddings on the
        full test dataset.}
    \vspace{-0.8em}
\end{figure*}

%% file: sections/method.tex
\section{Video Pose Distillation}
\label{sec:method}

Our strategy is to distill inaccurate pose estimates from an existing,
off-the-shelf pose detector --- the \emph{teacher} ---, trained on generic pose
datasets, into a --- \emph{student} --- network that is specialized to generate
robust pose descriptors for videos in a specific target sport domain
(\autoref{fig:system_diagram}).
The student (\autoref{sec:student}) takes RGB pixels and optical flow, cropped around the athlete, as input.
It produces a descriptor, from which we regress the athlete's pose as emitted by
the teacher (\autoref{sec:teacher}).
We run this distillation process over a large, \emph{uncut and unlabeled} corpus
of target domain videos (\autoref{sec:training_data}), using the sparse set of
high-confidence teacher outputs as weak supervision for the student.

Since the teacher is already trained, \OURMETHOD requires no new pose annotations in
the target video domain. Likewise, no downstream application-specific labels (e.g.,
action labels for recognition) are needed to learn pose features.
\OURMETHOD does, however, require that the athlete be identified in each input frame,
so we assume that an approximate bounding box for the athlete is provided in each frame as part of the dataset.
Refer to~\autoref{sec:discussion} for discussion and limitations.

\subsection{Teacher Network}
\label{sec:teacher}

To stress that \OURMETHOD is a general approach that can be applied to different teacher models, we propose two teacher variants of \OURMETHOD.
The first uses an off-the-shelf pose estimator~\cite{hrnet} to estimate 2D joint
positions from \Frame{\Time}, the RGB pixels of the \Time-th frame.
We normalize the 2D joint positions by rescaling and centering as in~\cite{prvipe},
and we collect the joint coordinates into a vector
$\WeakPose{\Time}\in\mathbb{R}^\PoseDim$.
We refer to this as 2D-\OURMETHOD since the teacher generates 2D joint
positions.

Our second teacher variant further processes the 2D joint positions into a
\emph{view-invariant} pose descriptor, emitted as $\WeakPose{\Time}$.
Our implementation uses \VIPESTAR to generate this descriptor.
\VIPESTAR is a reimplementation of concepts from  Pr-VIPE~\cite{prvipe} that is extended to
train on additional synthetic 3D pose data~\cite{amass,3dpeople,nba2k} for better
generalization.
We refer to this variation as VI-\OURMETHOD since the teacher generates a
view-invariant pose representation.
(See supplemental for details about \VIPESTAR and its quality compared to
Pr-VIPE.)

\subsection{Student Feature Extractor}
\label{sec:student}

Since understanding an athlete's motion, not just their current pose, is a key
aspect of many sports analysis tasks, we design a student feature extractor that
encodes information about both the athlete's current pose \WeakPose{\Time} and
the rate of change in pose $\PoseMotion{\Time}:=\WeakPose{\Time}-\WeakPose{\Time-1}$.

The student is a neural network \Student that consumes a color
video frame $\Frame{\Time}\in\mathbb{R}^{\ImageDims}$, cropped around the
athlete, along with its optical flow
$\Flow{\Time}\in\mathbb{R}^{\FlowDims}$, from the previous frame. \Height and \Width
are the crop's spatial dimensions, and \Time denotes the frame index.
The student produces a descriptor
$\Student\left(\Frame{\Time},\Flow{\Time}\right)\in\mathbb{R}^\EmbeddingDim$,
with the same dimension \EmbeddingDim as the teacher's output.
We implement \Student as a standard ResNet-34~\cite{resnet} with 5 input channels, and we resize the input
crops to $128\times128$.

During distillation, the features emitted by \Student are passed through an
auxiliary decoder \PoseDecoder, which predicts \emph{both} the current pose
\WeakPose{\Time} and the temporal derivative \PoseMotion{\Time}.
Exploiting the temporal aspect of video, \PoseMotion{\Time} provides an additional
supervision signal that forces our descriptor to capture motion in addition to the current pose.
\PoseDecoder is implemented as a fully-connected network, and
we train the combined student pathway
$\PoseDecoder \circ \Student$
using the following objective:
\begin{equation}
    \minimize_{\Student,\PoseDecoder}
        \sum_{\Time=1}^\NumFrames {
            \begin{Vmatrix}
                D\left(
                    F\left(\Frame{\Time},\Flow{\Time}\right)
                \right) -
                \begin{bmatrix}
                    \WeakPose{\Time} \\
                    \PoseMotion{\Time}
                \end{bmatrix}
            \end{Vmatrix}
        }^2_2
\end{equation}
Since only \Student is needed to produce descriptors during inference,
we discard \PoseDecoder at the end of training.

Unlike its teacher, which was trained to recognize a general distribution of poses
and human appearances, the student~\Student \emph{specializes} to frames and optical flow
in the new target domain (e.g., players in tennis courts).
Specialization via distillation allows \Student to focus on patterns present in the sports data that explain pose.
We do not expect, nor do downstream tasks require, that \Student encode poses or people not seen in the target
domain (e.g., sitting on a bench, ballet dancers), although they may be part of the teacher's training distribution.
Experiments in~\autoref{sec:eval} show that our pose descriptors,
$\Student(\Frame{\Time},\Flow{\Time})$, improve accuracy on
several applications, including few-shot, fine-grained action recognition.

\subsection{Training Data Selection and Augmentation}
\label{sec:training_data}

\noindent\textbf{Data selection.}
The teacher's output may be noisy due to challenges such as motion blur and occlusion
or because of domain shift between our target videos and the data that the teacher was trained on.
To improve the student's ability to learn and to discourage memorization of the teacher's noise,
we exclude frames with low pose confidence scores (specifically, {\em mean estimated joint score}) from the teacher's weak-supervision set.
By default, the threshold is 0.5, although 0.7 is used for tennis. Tuning this threshold has an effect on the quality of the distilled features (see supplemental for details).
We also withhold a fixed fraction of frames (20\%) uniformly at random as a validation set for the student.

\heading{Data augmentation.}
We apply standard image augmentations techniques such as
random resizing and cropping; horizontal flipping; and color and noise jitter,
when training the student \Student.
To ensure that left-right body orientations are preserved when horizontally augmenting
\Frame{\Time} and \Flow{\Time}, we also must flip the teacher's output
\WeakPose{\Time}.
For 2D joint positions and 2D-\OURMETHOD, this is straightforward.
To flip \VIPESTAR (itself a chiral pose embedding) used to train VI-\OURMETHOD,
we must flip the 2D pose inputs to \VIPESTAR and then re-embed them.

%% file: sections/result_recognize.tex
\input{figures/few_shot_action}

\section{Results}
\label{sec:eval}

We evaluate the features produced by \OURMETHOD on four
fine-grained sports datasets that exhibit a wide range of motions.

\heading{Figure skating}
consists of 371 singles mens' and womens' short program performances from the
Winter Olympics (2010-18) and World Championships (2017-19), totalling 17 video
hours.
In the classification task, \fs, there are six jump types defined by the
ISU~\cite{isu}.
All videos from 2018 (134 routines, 520 jumps) are held out for
testing. The remaining jumps are split 743/183 for training/validation.

\heading{Tennis}
consists of nine singles matches from two tournaments
(Wimbledon and US Open), with swings annotated at the frame of ball contact~\cite{vid2player}.
There are seven swing classes in \tennis.
The training/validation sets contain 4,592/1,142
examples from five matches and the test set contains 2,509 from the remaining four matches.
Split by match video, this dataset is challenging due to the limited diversity in clothing and unique individuals (10 professional players).

\heading{Floor exercise.}
We use the womens' floor exercise event (\fx) of the FineGym99
dataset~\cite{finegym}, containing 1,214 routines (34 hours). There are 35
classes and 7,634 actions.

\heading{Diving48}~\cite{diving48}
contains 16,997 annotated instances of 48 dive sequences, defined by FINA~\cite{fina}.
We evaluate on the corrected V2 labels released by the authors and retrain the
existing state-of-the-art  method, GSM~\cite{gsm}, for comparison.

\vspace{0.5em}
All four datasets contain frames in which pose is not well
estimated or uncertain, though their distribution varies (see supplemental for details).
As mentioned beforehand, pose estimates are typically worse in frames with fast motion, due to motion
blur and atypical, athletic poses such as flips or dives; see~\autoref{fig:bad_poses} for examples.
A common challenge across these datasets, the fast-motion frames are
often necessary for discriminating the fine-grained actions of interest.

We assume the subject of the action is identified and tracked.
With multiple humans in the frame, fast-moving athletes in challenging poses are
often missed otherwise: i.e., detected at lower confidence than static audience members or judges, or not detected at all.
{\em For fair comparison, we boost the baselines by providing them the same
inputs as our method, which improves their results significantly.}

\input{tables/full_data_action}

\input{tables/ablations}

\subsection{Fine-Grained Action Recognition}
\label{sec:action_recognition}

Fine-grained action recognition tests \OURMETHOD's ability to capture precise details about an athlete's pose and motion.
We consider both the few-shot setting, where only a limited number of action examples are
provided, and the traditional full supervision setting, where all of the action
examples in the training set are available.

Our \OURMETHOD features are distilled over the training videos in the sports corpus, uncut and without labels.
To extract features on the test set, we use the fixed \OURMETHOD student \Student.
VI-\OURMETHOD and 2D-\OURMETHOD features maintain the same
dimensions \PoseDim, of their teachers: $\PoseDim=64$ for \VIPESTAR and
$\PoseDim=26$ for normalized 2D joints.
For Diving48, \VIPESTAR has $\PoseDim=128$ because we also extract pose embeddings on the vertically flipped poses and concatenate them.
This data augmentation is beneficial for
\VIPESTAR due to the often inverted nature of diving poses, which are less well represented in the out-of-domain 3D pose datasets that \VIPESTAR is trained on.

\heading{Action recognition model.}
To use \OURMETHOD for action recognition, we first represent each action as a sequence of pose features.
We then classify actions using a bidirectional Gated Recurrent Unit network (BiGRU)~\cite{rnnchapter} trained atop the (fixed) features produced by the student \Student.
Since our features are chiral and many actions can be performed with either left-right orientation, we embed both the regular and horizontally
flipped frames with the student.
See supplemental for implementation details.

Prior pose embedding work has explored using sequence alignment followed by nearest-neighbor retrieval~\cite{prvipe}.
We also tested a nearest-neighbor search (NNS) approach that
uses dynamic time warping to compute a matching cost between sequences of pose features.
For NNS, each test example is searched against all the training
examples, and the label of the best aligned match is predicted.
The BiGRU is superior in most settings, though NNS can be effective in few-shot situations, and we indicate when this is the case.

\heading{Baselines.}
We compare our distilled 2D-\OURMETHOD and VI-\OURMETHOD features against several baselines.
\begin{enumerate}
    \item The {\em features from the teacher:}
        \VIPESTAR or the normalized 2D
        joint positions, using the same downstream action recognition models and data
        augmentations.
    \item {\em Skeleton-based:} a MS-G3D ensemble~\cite{msg3d} and
        ST-GCN~\cite{stgcn}. Both baselines receive the same tracked 2D poses used to supervise
        \OURMETHOD.
    \item {\em End-to-end:} GSM~\cite{gsm}, TSN~\cite{tsn}, and TRNms~\cite{trn} (multiscale).
        We test with both the cropped athletes and the full frame (w/o cropping) as inputs,
        and we find that cropping significantly improves accuracy in both the
        few-shot setting on all four datasets, and the full supervision
        setting on all datasets except for Diving48.
        When applicable, combined results with RGB and optical flow models are indicated as 2-stream.
\end{enumerate}

\subsubsection{Few-shot and limited supervision setting}
\label{sub:few_shot_action}

\noindent\textbf{Experiment protocol.}
Each model is presented $k$ examples of each action class
but may utilize unlabeled data or knowledge from other datasets as pre-training.
For example, skeleton-based methods rely on 2D pose detection; \VIPESTAR leverages out-of-domain 3D pose data; and
\OURMETHOD features are distilled on the uncut, unlabeled training videos.
This experimental setup mirrors real-world
situations where few labels are present but unlabeled and out-of-domain data are plentiful.
Our evaluation metric is top-1 accuracy on the full test set.
To control for variation in the training examples selected for each few-shot
experiment, we run each algorithm on five randomly sampled and fixed subsets of the data,
for each $k$, and report the mean accuracy.

\heading{Results.}
\autoref{fig:few_shot_action} compares 2D-\OURMETHOD and VI-\OURMETHOD features to their teachers (and other baselines).
On \fs and \tennis, VI-\OURMETHOD provides a slight improvement over its state-of-the-art teacher,~\VIPESTAR, with accuracies within a few percent.
\fx shows a large improvement and VI-\OURMETHOD increases accuracy by up to 10.5\% over \VIPESTAR at $k\leq32$ and 5\% over the MS-G3D ensemble at $k=64$.
Likewise, on Diving48, where end-to-end GSM and 2-stream TSN are otherwise better than the non-\OURMETHOD pose-based methods, VI-\OURMETHOD improves accuracy by 6.8 to 22.8\%.
Our results on \fx and Diving48 suggest that VI-\OURMETHOD helps to transfer the benefits of pose to datasets where it is most unreliable.

While view-invariant (VI) features generally perform better than their 2D analogues, the difference in accuracy between VI-\OURMETHOD and 2D-\OURMETHOD is more noticeable in sports with diverse camera angles (such as figure skating and floor exercise) and at small $k$, where
the action recognition model can only observe a few views during training.

\input{tables/retrieval}

\subsubsection{Traditional, full training set setting}
\label{sec:many_shot}

\OURMETHOD features are competitive even in the high-data regime (\autoref{tab:full_action_dataset}).
On all four datasets, both VI-\OURMETHOD and 2D-\OURMETHOD significantly improve accuracy over their teachers.
VI-\OURMETHOD also achieves state-of-the-art accuracy on the \fs (0.6\% over \VIPESTAR), \tennis (1.5\% over \VIPESTAR), and \fx (1.0\% over GSM, with cropped inputs) datasets.

Diving48 is especially challenging for pose-based
methods, and VI-\OURMETHOD performs worse than GSM, without cropping, by 1.6\%.
GSM, with cropping, is also worse by 1.5\%, possibly due to errors and limitations of our tracking.
VI-\OURMETHOD does, however, perform significantly better than the top pose-based baseline (8.4\% over MS-G3D, ensemble).

Our results demonstrate that \OURMETHOD's success is not limited to few-shot regimes. However, because many methods in~\autoref{tab:full_action_dataset} can produce high accuracies, at or above 90\%, when given ample data, we view improving label efficiency as a more important goal for \OURMETHOD and future work.

\subsubsection{Ablations and additional experiments}
\label{sub:ablations}

We highlight two important ablations of \OURMETHOD to understand the source of \OURMETHOD's improvements: (1) analyzing parts of the
distillation method and (2) distilling with only the action segments of the
video. We also consider (3) an unlabeled setting where \OURMETHOD is distilled over the entire video corpus.
Please refer to supplemental for additional experiments.

\heading{Analysis of the distillation method.}
\autoref{tab:ablate_table}(a) shows the increase in accuracy on action
recognition for ablated 2D-\OURMETHOD and VI-\OURMETHOD features when we distill without flow input $\Flow{\Time}$ and without motion prediction\footnote{The student mimics the teacher's $\WeakPose{\Time}$ output directly, without the auxiliary decoder \PoseDecoder and $\PoseMotion{\Time}$ in the training loss.}.
The incremental improvements are typically most pronounced in the few-shot setting, on the \fx and Diving48 datasets, where \OURMETHOD produces the largest benefits (see \autoref{sub:few_shot_action}).

With \VIPESTAR as the teacher, distillation alone from RGB can have a large effect (2.7\% and 7.7\%, at full and 16-shot settings on \fx; 7.9\% and 19.9\% on Diving48).
Adding flow in addition to RGB, without motion, gives mixed results. Finally, adding motion prediction and decoder \PoseDecoder,
further improves results (1.1\% and 1.5\% on \fx, at full and 16-shot; 2.1\% and 3.9\% on Diving48).
The effect of distilling motion on \fs and \tennis is mixed at the 16-shot setting, though the full  setting shows improvement.

2D-\OURMETHOD can be seen as an ablation of view-invariance (\VIPESTAR) and shows a similar pattern when further ablated.

\heading{Training \OURMETHOD on action parts of video only.}
Fine-grained action classes represent less than 7\%, 8\%, and 28\% of the video
in \fs, \fx, and \tennis.
We evaluate whether distillation of VI-\OURMETHOD over uncut video improves
generalization on action recognition, by distilling VI-\OURMETHOD features with
{\em only the action parts} of the training videos.

The results are summarized in~\autoref{tab:ablate_table}(b) and
show that distilling with only the action video performs worse on our
datasets.
This is promising because (1) uncut performances are much easier to obtain than
performances with actions detected, and (2) in the low-supervision setting,
VI-\OURMETHOD improves accuracy even if actions have not been detected in the
rest of the training corpus.
This also suggests that distilling over more video improves the quality of the features.

\heading{Distillation with the entire video corpus.}
An unlabeled corpus is often the starting point
when building real-world applications with videos in a new domain (e.g., \cite{vid2player}).
Because \OURMETHOD is supervised only by machine-generated
pose estimates from unlabeled video, \OURMETHOD features can be distilled over all of the video available, not just the training data.\footnote{This setting is similar to \cite{selfdomainshift,tttrain}, which propose self-supervision to align the training and testing distributions in situations with large domain shift.}
\autoref{tab:ablate_table}(c) shows results when VI-\OURMETHOD is
distilled jointly with both the training and testing videos, {\em \mbox{uncut} and \mbox{without labels}}.
The improvement, if any, is minor on all four datasets ($\leq$1.5\%, attained on \tennis at 16-shot) and demonstrates that VI-\OURMETHOD, distilled over a large dataset, is able to generalize without seeing the test videos.

%% file: figures/few_shot_action.tex
\begin{figure*}[thp]
    \centering
    \includegraphics[width=\textwidth]{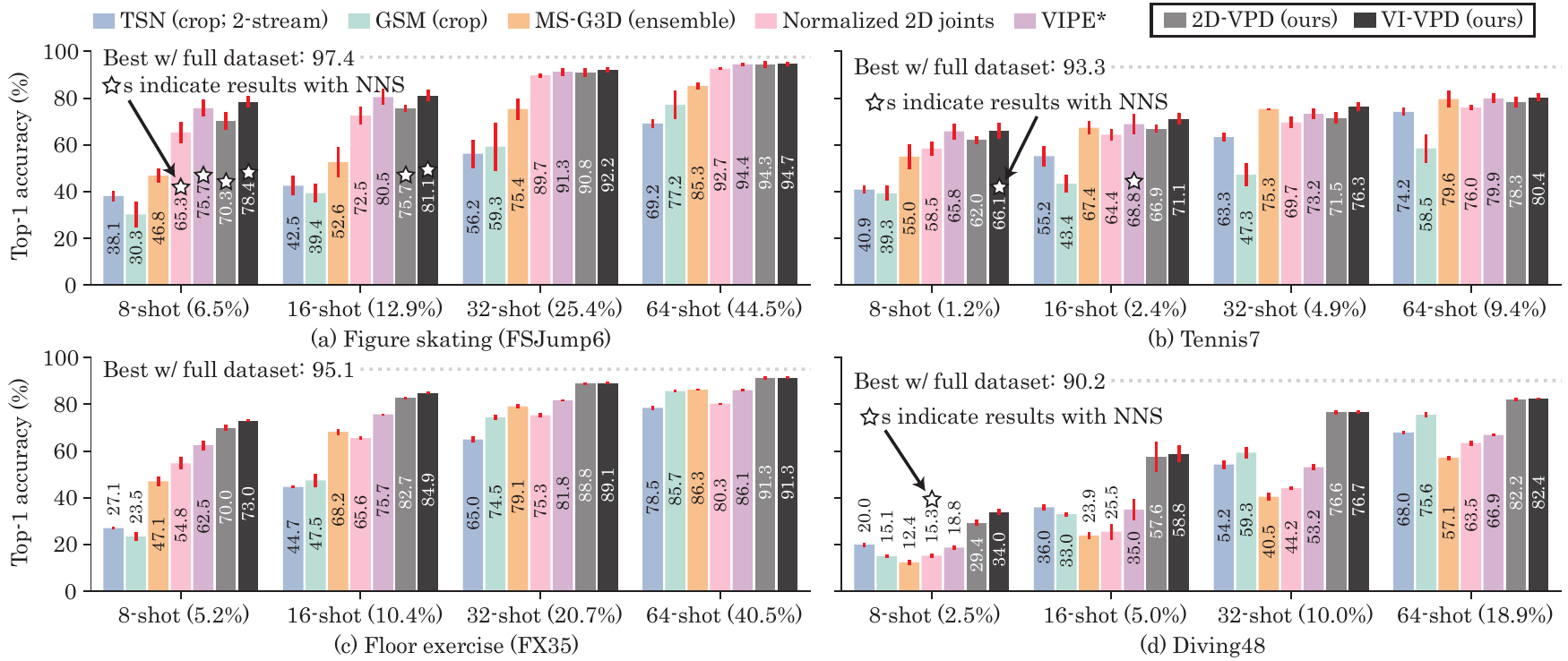}
    \vspace{-0.27in}
    \caption{{\bf Accuracy on few-shot fine-grained action recognition.} Percentages give the fraction of the full training set.
    State-of-the-art accuracies using the full dataset as supervision are indicated as a dashed line for reference (see~\autoref{tab:full_action_dataset}).
    Pose-based baselines (MS-G3D~\cite{msg3d}, 2D joints, and \VIPESTAR) surpass end-to-end baselines (GSM~\cite{gsm} and TSN~\cite{tsn}) in few-shot settings on every dataset except Diving48, demonstrating both the importance of pose when labels are scarce and the challenge when pose is unreliable.
    VI-\OURMETHOD significantly outperforms the baselines and prior methods on \fx and Diving48;
    accuracy on \fs and \tennis also improves slightly, but remains similar to \VIPESTAR.
    {\em Starred results above use nearest-neighbor search (NNS) instead of the BiGRU architecture (NNS performed better in these cases; see supplemental for full results)}.
    }
    \label{fig:few_shot_action}
    \vspace{-0.8em}
\end{figure*}

%% file: tables/full_data_action.tex
\begin{table}[t!]
    \centering
    \begin{tabularx}{\columnwidth}{lrrrr}
        \toprule
        \multicolumn{1}{l}{\hspace{7pt}Dataset (Top-1 acc)}
            & \multicolumn{1}{r}{\hspace{-2em}\rotatebox[origin=r]{-22}{\hspace{4pt}\fs\hspace{-4pt}}}
            & \multicolumn{1}{r}{\hspace{-2em}\rotatebox[origin=r]{-22}{\tennis\hspace{-4pt}}}
            & \multicolumn{1}{r}{\hspace{-2em}\rotatebox[origin=r]{-22}{\fx\hspace{-4pt}}}
            & \multicolumn{1}{r}{\hspace{-2em}\rotatebox[origin=r]{-22}{Diving48\hspace{-4pt}}} \\
        \midrule
        Random      & 16.7 & 14.3 & 2.9 & 2.1 \\
        Top class   & 33.7 & 46.7 & 7.5 & 8.3 \\
        \midrule
        \multicolumn{5}{l}{\hspace{7pt}End-to-end} \\
        \midrule
        \multicolumn{2}{l}{\textsuperdagger TRNms (2-stream)~\cite{finegym,trn}}
                && 84.9 & \\
        \textsuperdagger TimeSformer-L~\cite{timesformer} &
                &&& 81.0 \\
        TSN~\cite{tsn} (w/o crop) &
                57.9 & \multicolumn{1}{c}{-} & 83.2 & 82.3 \\
        TSN (crop) &
                81.2 & 87.8 & 88.5 & 83.6 \\
        TSN (crop; 2-stream) &
                82.7 & \underline{90.9} & 90.4 & 83.6 \\
        TRNms~\cite{trn} (w/o crop) &
                68.7 & \multicolumn{1}{c}{-} & 81.5 & 80.5 \\
        TRNms (crop) &
                77.7 & 55.5 & 87.1 & 81.8 \\
        TRNms (crop; 2-stream) &
                84.0 & 76.3 & 87.3 & 81.5 \\
        GSM~\cite{gsm} (w/o crop) &
                42.1 & \multicolumn{1}{c}{-} & 90.3 & \underline{\textbf{90.2}} \\
        GSM (crop) &
                \underline{90.6} & 67.1 & \underline{93.6} & 88.7\\
        \midrule
        \multicolumn{5}{l}{\hspace{7pt}Skeleton / pose-based (w/ tracked 2D poses)} \\
        \midrule
        \multicolumn{2}{l}{\textsuperdagger ST-GCN (w/o tracking)~\cite{finegym,stgcn}}
               & & 40.1 & \\
        ST-GCN~\cite{stgcn} &
                88.7 & 88.4 & 80.3 & 64.8 \\
        MS-G3D (ensemble)~\cite{msg3d} &
                \underline{91.7} & \underline{91.0} & \underline{92.1} & \underline{80.2} \\
        \midrule
        \multicolumn{5}{l}{\hspace{7pt}Pose features (w/ BiGRU)} \\
        \midrule
        Normalized 2D joints &
                95.5 & 90.9 & 86.9 & 75.7 \\
        {\bf (Ours)} 2D-\OURMETHOD &
                97.0 & 92.6 & 94.5 & 86.4 \\
        \VIPESTAR &
                96.8 & 91.8 & 90.8 & 78.6 \\
        {\bf (Ours)} VI-\OURMETHOD &
                \underline{\textbf{97.4}} &
                \underline{\textbf{93.3}} &
                94.6 &
                88.6 \\
        {\bf (Ours)} Concat-\OURMETHOD &
                96.2 & 93.2 & \underline{\textbf{95.1}} & \underline{88.7} \\
        \bottomrule
    \end{tabularx}
    \vspace{-0.08in}
    \caption{{\bf Accuracy on fine-grained action recognition with all of the training data.}
        Top results overall are \textbf{bolded} and per method category are \underline{underlined}. \textdagger~indicates best results from prior work.
        VI-\OURMETHOD achieves SOTA accuracy on \fs, \tennis, and
        \fx, even when baselines are improved with tracking and cropped inputs. On
        Diving48, VI-\OURMETHOD trails end-to-end GSM (w/o crop) by 1.6\%.
        VI-\OURMETHOD and 2D-\OURMETHOD features can both be competitive; concatenating them (Concat-\OURMETHOD) may improve accuracy slightly.
    }
    \label{tab:full_action_dataset}
    \vspace{-0.8em}
\end{table}

%% file: tables/ablations.tex
\begin{table*}[t]
    \centering
    \begin{tabularx}{\textwidth}{clccccccccccc}
        \toprule
        \multicolumn{2}{c}{Dataset}
                & \multicolumn{2}{c}{\fs}       &
                & \multicolumn{2}{c}{\tennis}   &
                & \multicolumn{2}{c}{\fx}       &
                & \multicolumn{2}{c}{Diving48} \\
        \multicolumn{2}{c}{Input features \textbackslash~Amount of training data}
                & Full & \multicolumn{1}{c}{16-shot}       &
                & Full & \multicolumn{1}{c}{16}       &
                & Full & \multicolumn{1}{c}{16}       &
                & Full & \multicolumn{1}{c}{16} \\
        \midrule
        (a)\hspace{-0.1in}
         & Normalized 2D joints (teacher)
                & 95.5 & 72.5  &
                & 90.9 & 64.3                   &
                & 86.9 & 65.6                   &
                & 75.7 & 25.5 \\
        & \hspace{8pt} distilled w/o motion; RGB
                & 96.1 & 73.2  &
                & 90.9 & 66.5                   &
                & 92.0 & 76.3                   &
                & 85.3 & 52.8 \\
        & \hspace{8pt} distilled w/o motion; RGB \& flow
                & 95.8 & 74.6  &
                & 91.7 & 67.0                   &
                & 91.6 & 76.6                   &
                & 85.6 & 53.0 \\
        & \hspace{8pt} 2D-\OURMETHOD: distilled w/ motion; RGB \& flow\vspace{0.25em}
                & 97.0 & 74.4  &
                & 92.6 & 66.9                   &
                & 94.5 & 82.7                   &
                & 86.4 & 57.6 \\

        & \VIPESTAR (teacher)
                & 96.8 & 80.5  &
                & 91.8 & 67.0                   &
                & 90.8 & 75.7                   &
                & 78.6 & 35.0 \\
        & \hspace{8pt} distilled w/o motion; RGB
                & 97.1 & \textbf{81.3}  &
                & 92.1 & 67.6                   &
                & 93.5 & 83.4                   &
                & 86.5 & 54.9 \\
        & \hspace{8pt} distilled w/o motion; RGB \& flow
                & 97.3 & 79.3  &
                & 91.7 & 69.7                   &
                & 92.9 & 83.2                   &
                & 85.9 & 53.7 \\
        & \hspace{8pt} VI-\OURMETHOD: distilled w/ motion; RGB \& flow
                & \textbf{97.4} & 80.2  &
                & \textbf{93.3} & \textbf{71.1}                  &
                & \textbf{94.6} & \textbf{84.9}                   &
                & \textbf{88.6} & \textbf{58.8} \\
        \midrule
        (b)\hspace{-0.1in}
        & VI-\OURMETHOD (distilled on action video only)
                & 96.3 & 79.4  &
                & 92.4 & 69.1                   &
                & 94.1 & 84.3                   &
                & - & - \\
        \midrule
        (c)\hspace{-0.1in}
        & VI-\OURMETHOD (distilled w/ the entire video corpus)
                & 97.2 & 81.9  &
                & 93.8 & 72.6                   &
                & 94.5 & 84.9                   &
                & 88.4 & 59.6 \\
        \bottomrule
    \end{tabularx}
    \vspace{-0.08in}
    \caption{{\bf Action recognition experiments.}
        Top-1 accuracy in the full training set and
        16-shot scenarios with (a) ablations to the distillation methodology,
        (b) when only the action parts of the dataset are used for distillation,
        and (c) when VI-\OURMETHOD features are distilled over the entire video corpus
        (including the testing videos, without labels).
        Results are with the BiGRU.
    }
    \label{tab:ablate_table}
    \vspace{-0.8em}
\end{table*}

%% file: tables/retrieval.tex
\begin{table*}[tp]
    \centering
    \begin{tabularx}{\textwidth}{lccccccccccccccc}
        \toprule
        \multicolumn{1}{c}{Dataset}
                & \multicolumn{3}{c}{\fs}       &
                & \multicolumn{3}{c}{\tennis}   &
                & \multicolumn{3}{c}{\fx}       &
                & \multicolumn{3}{c}{Diving48} \\
        \multicolumn{1}{c}{k}
                & 1 & 10 & \multicolumn{1}{c}{50} &
                & 1 & 10 & \multicolumn{1}{c}{50} &
                & 1 & 10 & \multicolumn{1}{c}{50} &
                & 1 & 10 & 50 \\
        \midrule
        Normalized 2D joints\hspace{7pt}
                & 91.8 & 84.8 & 73.8 &
                & 91.8 & 88.1 & 82.1 &
                & 71.6 & 57.4 & 39.0 &
                & 34.5 & 22.1 & 14.6 \\
        {\bf (Ours)} 2D-\OURMETHOD
                & 92.5 & 86.4 & 76.3 &
                & {\bf 93.1} & 90.0 & 84.6 &
                & 79.7 & 66.8 & 47.5 &
                & {\bf 64.4} & {\bf 43.8} & {\bf 27.9} \\
        \VIPESTAR
                & 92.9 & 85.1 & 75.7 &
                & 92.4 & 90.0 & 85.9 &
                & 72.2 & 60.1 & 46.6 &
                & 36.1 & 24.1 & 15.1 \\
        {\bf (Ours)} VI-\OURMETHOD
                & {\bf 93.6} & {\bf 86.8} & {\bf 78.0} &
                & 92.8 & {\bf 90.6} & {\bf 86.3} &
                & {\bf 80.8} & {\bf 68.6} & {\bf 52.4} &
                & 60.9 & 40.9 & 25.4 \\
        \bottomrule
    \end{tabularx}
    \vspace{-0.08in}
    \caption{{\bf Action retrieval: Precision@k results (\%) ranked by alignment score with dynamic time warping.} \OURMETHOD leads to more relevant results on all four datasets. Gains on \fs and \tennis are modest, while the large improvements on \fx and Diving48 suggest that \OURMETHOD features are superior in cases when pose estimates are the most unreliable.}
    \label{tab:action_retrieval}
    \vspace{-0.5em}
\end{table*}

%% file: sections/result_other.tex
\input{tables/detection}

\subsection{Action Retrieval}
\label{sec:eval:actionretrieval}

Action retrieval measures how well \OURMETHOD features can be used to search for similar unlabeled actions. Here, the \OURMETHOD features are distilled on the entire unlabeled corpus.

\heading{Experiment protocol.}
Given a query action, represented as a sequence of pose features, we rank all other actions in the corpus using the $L_2$ distance between pose features and dynamic time warping to compute an alignment score.
A result is considered relevant if it has the same fine-grained action label as the query, and we assess relevance by the precision at \emph{k} results, averaged across all the queries.

\heading{Results.}
At all cut-offs in~\autoref{tab:action_retrieval} and in all four datasets, \OURMETHOD features outperform their teachers.
Sizeable improvements are seen on \fx and Diving48.
View-invariance does not always result in the highest precision if the number of camera angles is limited (e.g., \tennis and Diving48), though it may be helpful for retrieving more diverse results.

\subsection{Pose Features for Few-Shot Action Detection}
\label{sec:eval:detection}

Detection of fine-grained actions, at fine temporal granularity and with few
labels, enables tasks such as few-shot recognition and retrieval.
We evaluate \OURMETHOD features on the figure skating and tennis datasets, to temporally
localize the jumps and the swings, respectively.
The average jump is 1.6 seconds in length ($\approx$40 frames), while
a swing is defined to be the 200 ms around the frame of ball contact ($\approx$5 frames).

\heading{Experiment protocol.}
{\em We follow the same video-level train/test splits as \fs and \tennis, and distill features on the training videos only.}
As a simple baseline method, we train a BiGRU that outputs per-frame predictions, which are merged to
produce predicted action intervals (see supplemental for details).
The BiGRU is trained on ground-truth temporal labels from five routines (figure skating)
and five points (tennis).
For more consistent results, we perform five-fold cross-validation and ensemble the per-frame predictions.
In \autoref{tab:detection}, we report average precision (AP) at various levels of temporal intersection over union (tIoU).

\heading{Results.}
\OURMETHOD improves AP on both tasks.
The short duration of tennis swings means that noise in per-frame pose estimates has a large impact, and
VI-\OURMETHOD improves AP at every tIoU threshold (up to 7.4 over \VIPESTAR at $\text{tIoU}=0.5$).

%% file: tables/detection.tex
\begin{table}[t]
    \centering
    \begin{tabularx}{\columnwidth}{lcrrrrr}
        \toprule
        \multicolumn{1}{c}{Temporal IoU}
            & \multicolumn{1}{c}{0.3} & \multicolumn{1}{c}{0.4} & \multicolumn{1}{c}{0.5}
            & \multicolumn{1}{c}{0.6} & \multicolumn{1}{c}{0.7} \\
        \midrule
        \multicolumn{6}{l}{Figure skating jumps (trained on five routines)} \\
        \midrule
        \hspace{-7pt} Pretrained R3D~\cite{resnet3d}
            & 39.5 & 30.0 & 23.1 & 15.0 & 9.0  \\
        \hspace{-7pt} Normalized 2D joints
            & 80.6 & 70.0 & 53.5 & 40.2 & 24.6 \\
        \hspace{-7pt} {\bf (Ours)} 2D-\OURMETHOD
            & 85.7 & 77.8 & \textbf{61.5} & 47.6 & 25.8 \\
        \hspace{-7pt} \VIPESTAR
            & 84.5 & 76.8 & 59.3 & 45.3 & 26.7 \\
        \hspace{-7pt} {\bf (Ours)} VI-\OURMETHOD
            & \textbf{86.1} & \textbf{78.6} & 60.7 & \textbf{47.9} & \textbf{28.7} \\
        \midrule
        \multicolumn{6}{l}{Tennis swings at 200 ms (trained on five points)} \\
        \midrule
        \hspace{-7pt} Pretrained R3D~\cite{resnet3d}
            & 41.3 & 37.8 & 29.9 & 15.8 & 7.6 \\
        \hspace{-7pt} Normalized 2D joints
            & 59.7 & 58.2 & 43.7 & 24.6 & 10.3 \\
        \hspace{-6pt} {\bf (Ours)} 2D-\OURMETHOD
            & 67.4 & 66.5 & 54.0 & 28.4 & 13.1 \\
        \hspace{-7pt} \VIPESTAR
            & 67.4 & 65.8 & 51.2 & 28.9 & 12.3 \\
        \hspace{-7pt} {\bf (Ours)} VI-\OURMETHOD
            & \textbf{73.5} & \textbf{72.6} & \textbf{58.6} & \textbf{32.9} & \textbf{13.8} \\
        \bottomrule
    \end{tabularx}
    \vspace{-0.08in}
    \caption{{\bf Few-shot action detection: Average precision (AP) at various levels of temporal IoU.} VI-\OURMETHOD features improve AP over \VIPESTAR and the other baselines.}
    \label{tab:detection}
    \vspace{-0.8em}
\end{table}

%% file: sections/discussion_and_conclusion.tex
\section{Limitations and Discussion}
\label{sec:discussion}

\noindent\textbf{Subject tracking} is needed for \OURMETHOD to ensure that the pose is of the correct person. Real-world sports video often contains many people, such as audience and judges, in addition to the subject. The tracking annotations in the datasets in~\autoref{sec:action_recognition} are computed automatically using off-the-shelf models and heuristics (see supplemental for details). This is possible because athletes are salient in appearance, space, and time --- sports video is a natural application for work on tracking~\cite{sort,deepsort} and detecting salient regions~\cite{saliencymodellooknext}.
We observe that the difference in accuracy between the tracked and non-tracked inputs on other prior methods such as~\cite{gsm,tsn,stgcn} can be staggering (48\% on \fs for GSM~\cite{gsm} and 40\% on \fx for ST-GCN~\cite{stgcn}; see~\autoref{tab:full_action_dataset}).

To evaluate the quality of our pose features, we focused on motion by a single athlete or synchronized athletes (contained in Diving48). Tasks and actions involving many people require a more sophisticated downstream model that can handle multiple descriptors or poses per frame.

\heading{Future work.} 
First, the 2D pose estimates used to supervise \OURMETHOD are inherently ambiguous with respect to camera view, and additional information such as depth or a behavioral prior could help alleviate this ambiguity. Other weak supervision sources, in addition to motion and VIPE, may also help.
Second, our distillation process is offline; supporting online training, similar to~\cite{jitnet,tttrain}, at the pose feature extraction stage could be beneficial in time-evolving datasets.
Distillation for explicit 2D or 3D pose estimation is another possibility.
Although \OURMETHOD features can improve accuracy with limited data, performance on few-shot and semi-supervised tasks still has much room to improve, and we hope that future work continues to explore these  topics.

\section{Conclusion}
\label{sec:conclusion}

Pose features are useful for studying human-centric action in novel sports video datasets.
However, such datasets are often challenging for off-the-shelf models. 
Our method, \OURMETHOD, improves the reliability of pose features in difficult and label-poor settings, 
by distilling knowledge from existing pose estimators. \OURMETHOD learns features that improve accuracy on both traditional and few-shot action understanding tasks in the target (sport) domain.
We believe that our distillation-based method is a useful paradigm for addressing challenges faced by applications in new video domains.

%% file: sections/ack.tex
\heading{Acknowledgements.}
This work is supported by the National Science Foundation (NSF) under III-1908727 and Adobe Research. We also thank the anonymous reviewers.

%% file: supplemental/implementation.tex
\section{Implementation: Video Pose Distillation}
\label{sec:impl:vpd}

This section provides additional implementation details for our method described in~\autoref{sec:method}.

\heading{Pose: $\WeakPose{\Time}$ definition.}
\OURMETHOD is not dependent on a specific 2D pose estimator or joint definition. We use an off-the-shelf HRNet~\cite{hrnet} to estimate pose in the detected region of the athlete, as is typical for top-down pose estimation. Heuristic tracking, described in~\autoref{sec:dataset_details}, can often provide bounding boxes in frames where person detection fails. We use only 13 of the 17 COCO~\cite{coco} keypoints (ignoring LEye, REye, LEar, and REar), and we apply the same joint normalization procedure as in~\cite{prvipe}.

\heading{Student inputs.}
The RGB crops \Frame{\Time} are derived from the spatial bounding boxes of the athlete in frame \Time.
We expand the bounding box to a square and then pad each side by 10\% or 25 pixels, whichever is greater.

Optical flow \Flow{\Time} is computed using RAFT~\cite{raft} between \Frame{\Time} and \Frame{\Time-1}, where
we crop the same location as \Frame{\Time} in the previous frame for \Frame{\Time-1}. In datasets where the frame rate differs between videos, a target frame rate of 25 frames per second (fps) determines \Frame{\Time-1}.
To obtain the final \Flow{\Time}, we subtract the median of the RAFT output, clip to $\pm20$ pixels, and quantize into 8-bits.

During training and inference, $\Frame{\Time}$ is scaled to a range of $\pm1$ and standardized with respect to the dataset RGB mean and standard deviation;
\Flow{\Time} is also centered to $\pm0.5$.
In video frames where the athlete was explicitly detected by Mask R-CNN with a score above 0.8 (see~\autoref{sec:dataset_details}), we use the predicted mask to jitter the background with Gaussian noise ($\sigma=0.05$) as data augmentation.

For performance reasons, we pre-compute \WeakPose{\Time}, \Frame{\Time}, and \Flow{\Time} in an offline manner for the entire corpus.

\heading{Auxiliary decoder \PoseDecoder} is a standard fully connected network, whose sole purpose is to provide supervision for training the student \Student. We use two hidden layers, each with dimension of 128. Note that the ablations without motion in~\autoref{tab:ablate_table} do not use \PoseDecoder and directly optimize $L_2$ loss between the student's output and the teacher's $\WeakPose{\Time}$.

\heading{Student training.}
The student is initialized with random weights.
In each training epoch, we randomly sample 20,000 frames \Time that meet the pose selection criteria outlined in~\autoref{sub:data_selection}.
We use an AdamW~\cite{adamw} optimizer with learning rate $5e^{-4}$ and a batch size of 100.
The student is trained for 1,000 epochs, though in practice the model often converges sooner and using a higher learning rate is also possible.
We use the loss on the held-out validation frames to select the best epoch.
On a single Titan V GPU, the student model trains in approximately 8 hours.

\section{Implementation: Action Recognition}

This section provides details about our fine-grained action recognition models and baselines.

\subsection{BiGRU Architecture}
\label{sub:gru}

This is a standard bidirectional-GRU~\cite{rnnchapter} architecture.
The model is trained on sequences of VI-\OURMETHOD, 2D-\OURMETHOD, \VIPESTAR, and normalized 2D joint position features.

\heading{The inputs} are variable length sequences of per-frame pose features (for each action). The features are sampled to 25 fps in \fx and Diving48, where frame rate varies from 25 to 60 fps. \fs is a small dataset and normalizing the features also reduces overfitting.

\heading{Architecture.} We use a two-layer BiGRU as the backbone, with a hidden dimension $h=128$.
The output of the BiGRU is a sequence $H\in \mathbb{R}^{2h \times t}$ of hidden states from the final layer.
To obtain a fixed size encoding of this sequence, we max-pool across the time steps in $H$.
To output an action class, the pooled encoding is sent to a fully connected network consisting of BN-Dropout-FC-ReLU-BN-Dropout-FC, with the FC dimensions being $2h$ and the number of output classes.

\heading{Training.} We train the network with AdamW~\cite{adamw} and a batch size of 50 for 500 epochs (200 on Diving48 due to the larger dataset). Learning rate is initially set to $1e^{-3}$ and adjusted with a cosine schedule. Dropout rate is $0.5$ on the dense layers and $0.2$ on the input sequence. Data augmentation consists of the horizontally flipped input sequences.

On a single Titan V GPU, our model takes 7 minutes to train for \fs, 25 minutes for \tennis, 50 minutes for \fx, and 100 minutes for Diving48 over the full datasets.

\heading{Inference.} At inference time, we feed the input sequence and its horizontal flip to the model; sum the predictions; and output the top predicted class.

\input{figures/weak_pose}

\subsection{Nearest-Neighbor Search}
\label{sub:nns_dtw}

Our nearest-neighbor search (NNS) uses sequence alignment cost with dynamic time warping (DTW).

\heading{The inputs} are the same as in~\autoref{sub:gru}, but with each feature vector normalized to unit length.

\heading{Inference.} We treat the training set as an index. Alignment cost between two sequences of features, normalized by sequence length, is calculated using DTW with pairwise $L_2$ distance and the symmetricP2 step pattern~\cite{sakoe1978}. Combinations of the regular and horizontally flipped pose sequences in the testing set and training set are considered, with the lowest cost match returned.

Because the computational complexity of inference grows linearly with training set size, this method is unsuited for larger datasets with more examples or classes.
DTW is also sensitive to factors such as the precision of the temporal boundaries and the duration of the actions.

\subsection{Additional Baselines}

We evaluated ST-GCN~\cite{stgcn}, MS-G3D~\cite{msg3d}, multiscale TRN~\cite{trn}, and GSM~\cite{gsm} on our datasets using the reference implementations released by the authors.
For TSN~\cite{tsn}, we used the code from the authors of GSM~\cite{gsm}.
The GSM~\cite{gsm} codebase extends the TRN~\cite{trn} and TSN frameworks, and we backported ancillary improvements (e.g., learning rate schedule) to the TRN codebase for fairness.

\heading{Skeleton based.}
The inputs to ST-GCN and MS-G3D are the tracked 2D skeletons of only the identified athlete.
For MS-G3D, we trained both the bone and joint feature models and reported their ensemble accuracy. Ensemble accuracy exceeded the separate accuracies in all of our experiments.

\heading{End-to-end.}
We follow the best Diving48 configuration in the GSM~\cite{gsm} paper for the GSM, TSN, and TRNms baselines. This configuration uses 16 frames, compared to 3 to 7 in earlier work~\cite{trn}, and samples 2 clips at inference time.
As seen in benchmarks by the authors of~\cite{finegym}, additional frames are immensely beneficial for fine-grained action recognition tasks compared to coarse-grained tasks, where the class can often be guessed in a few frames from context~\cite{danceinmall,mimetics}.
The backbone for these baselines is an InceptionV3~\cite{inceptionv3}, initialized using pretrained weights.

When comparing to TSN and TRN with optical flow, we train using the same cropped flow images as \OURMETHOD, described in~\autoref{sec:impl:vpd}.
Flow and RGB model predictions are ensembled to obtain the 2-stream result.
Recent architectures that model temporal information in RGB, such as GSM, often perform as well as or better than earlier flow based work.

\section{Implementation: Action Retrieval}

The search algorithm for action retrieval is identical to nearest neighbor search described in~\autoref{sub:nns_dtw}, for action recognition, except that the pose sequence alignment scores are retained for ranking.

\heading{Query set.}
For \fs, \tennis, and \fx we evaluate with the entire corpus as queries. For the much larger Diving48 dataset, we use the 1,970 test videos as queries.

\section{Implementation: Action Detection}
\label{sec:impl:fewshotdetection}

We evaluated pose features for few-shot figure skating jump and tennis swing detection. Our method should be interpreted as a baseline approach to evaluate \OURMETHOD features, given the lack of prior literature on temporally fine-grained, few-shot video action detection, using pose features.
More sophisticated architectures for accomplishing tasks such as generating action proposals and refining boundaries are beyond the scope of this paper.

\heading{The inputs} are the uncut, per-frame pose feature sequences.
For figure skating, the sequences are entire, 160 second long, short programs.
ISU~\cite{isu} scoring rules require that each performance contains two individual jumps and a jump combination (two jumps).
For tennis, each point yields two pose sequences, one for each player.
The points sampled for training have at least five swings each per player.

For the ResNet-3D~\cite{resnet3d} baseline, we extracted features for each frame using a Kinetics-400~\cite{kinetics} pretrained model on the $128\times128$ subject crops, with a window of eight frames. A limitation of this baseline is that actions (e.g., tennis swings) can be shorter than the temporal window.

\heading{Architecture.} We use a two-layer BiGRU as the backbone with a hidden dimension $h=128$. The hidden states at each time step from the final GRU layer are sent to a fully connected network consisting of BN-Dropout-FC-ReLU-BN-Dropout-FC, with the FC dimensions being $2h$ and 2 (a binary label for whether the frame is part of an action).

\heading{Training.} The BiGRU is trained on randomly sampled sequences of 250 frames from the training set. We use a batch size of 100, $1e^4$ steps with the AdamW~\cite{adamw} optimizer, and a learning rate of $1e^{-3}$. We apply dropout rates of $0.5$ on the dense layers and $0.2$ on the input sequence.
Because only five examples are provided in this few-shot setting, we use five-fold cross validation to train an ensemble.

The reported results are an average of separate runs on five randomly sampled, fixed few-shot dataset splits.

\heading{Inference.} We apply the trained BiGRU ensemble to the uncut test videos to obtain averaged frame-level activations.
Consecutive activations above 0.2 are selected as proposals; the low threshold is due to the large class imbalance because actions represent only a small fraction of total time.
A minimum proposal length of three frames is required.
The mean action length in the training data was also used to expand or trim proposals that are too short (less than $0.67\times$) or too long (greater than $1.33\times$).

%% file: figures/weak_pose.tex
\begin{figure*}[p]
    \centering
    \includegraphics[width=\textwidth]{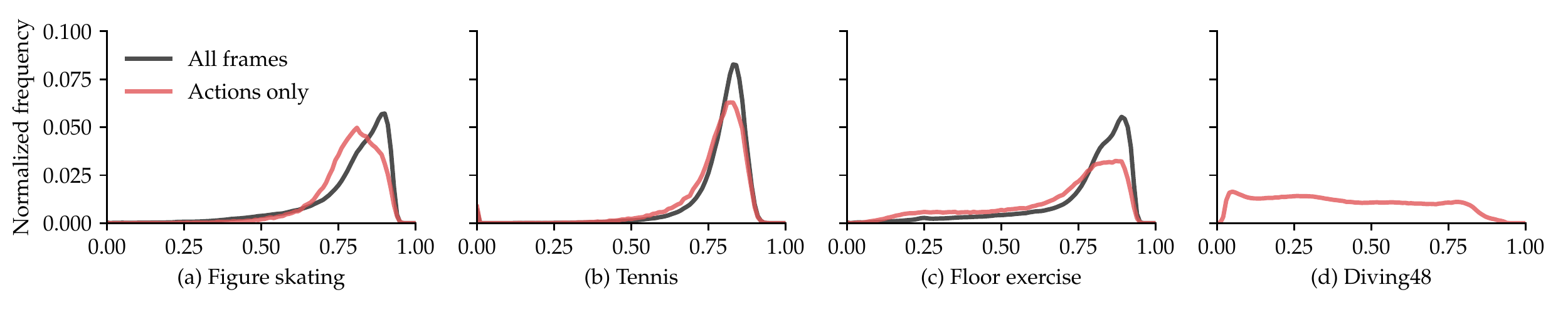}
    \vspace{-0.33in}
    \caption{
        {\bf Distribution of mean estimated joint scores in each dataset.}
        A flatter distribution with more mass to the left indicates greater uncertainty in the estimates.
        The distribution of joint scores produced by the pose estimator varies by dataset and whether the frames are part of actions or not.}
    \label{fig:pose_score}
    % \vspace{-1em}
\end{figure*}

%% file: supplemental/experiments.tex
\input{tables/weak_pose_stats}
\input{tables/nns_vs_gru_action}
\input{tables/detection_threshold}
\input{tables/other_action_arch}
\input{tables/gsm_crop_diving48}

\section{Additional Experiments}
\label{sec:extra_experiments}

This section includes results of additional ablations, analysis, and baselines omitted from the main text.

\subsection{Ablation: Data Selection Criterion}
\label{sub:data_selection}

{\em Mean estimated joint score} from the teacher pose estimator is used as the weak-pose selection criterion.
\autoref{fig:pose_score} shows the distribution of such scores in each of the four sports datasets.
Notice that the teacher produces significantly less confident pose estimates on the floor exercise (\fx) and Diving48 datasets, and also on the labeled action portions of all four datasets.

While the optimal selection threshold is ultimately dependent on the calibration and quality of the pose estimator used,
we evaluate the effect of tuning the weak-pose selection criterion on three of our datasets: \tennis, \fx, and Diving48. \autoref{tab:sparse_pose} shows results with VI-\OURMETHOD when various thresholds are applied.
There is benefit to ignoring the least confident pose estimates, though setting the threshold too high also diminishes performance, as insufficient data remains to supervise the student.

\subsection{Ablation: NNS vs. BiGRU for Recognition}

\autoref{fig:few_shot_action} notes that the BiGRU classifier for action recognition generally performed better than NNS, except in extremely data-scarce settings, where there are simultaneously few classes and examples per class.
\autoref{tab:fs_nns_dtw} presents results for both the BiGRU and NNS.

\subsection{Ablation: Activation Threshold for Detection}

In~\autoref{sec:impl:fewshotdetection}, we use a frame-level activation threshold of 0.2 when proposing action intervals for few-shot action detection.
\autoref{tab:detection_threshold} shows the impact on average precision (AP) of other thresholds, scored at 0.5 temporal intersection over union (tIoU).
The results are similar at nearby thresholds and results at 0.2 are reported for consistency.

\subsection{Ablation: Action Recognition Architectures}

The BiGRU described in~\autoref{sub:gru} was used in our experiments for consistency.
This section includes a number of additional simple, well-studied architectures that we also tested.
Results from these models are given in~\autoref{tab:other_action_arch} and are often similar; the BiGRU is not necessarily the best performing model in all situations.
As~\autoref{sec:action_recognition} shows, however, the BiGRU is competitive with recent, state-of-the-art methods when trained with \VIPESTAR or our VI-\OURMETHOD features.

\subsection{Baseline: GSM Without Cropping on Diving48}

In~\autoref{sub:few_shot_action}, on few-shot action recognition, we reported results from GSM~\cite{gsm} with cropping.
This is despite GSM, without cropping, having higher accuracy in the full supervision setting on Diving48~\cite{diving48} (see~\autoref{tab:full_action_dataset}).
\autoref{tab:gsm_few_shot} shows that GSM, with cropping, is the stronger baseline when limited supervision is available.

We speculate that cropping forces the GSM model focus on the diver in few-shot settings. In the full supervision setting, the GSM model can learn this information by itself and is limited by noise in the crops and the loss of other information from the frame (e.g., the other diver in synchronized diving; the 3 metre springboard or 10 metre platform; and spatial information).

\subsection{Analysis: Visualizing Distilled 2D Pose}
\label{sub:visualize_distilled_2d}

Although the goal of this paper is to distill pose features for downstream tasks, this section provides preliminary qualitative results on how well distilled features mimic their teachers and reflect the explicit 2D pose.
Because the learned \VIPESTAR and \OURMETHOD features are not designed to be human interpretable, we use normalized 2D joint positions (described in~\autoref{sec:impl:vpd}) as the teacher instead, and we train an ablated student without the auxiliary decoder for motion.

\autoref{fig:pose_examples} compares the teacher's normalized 2D joint features to the student's distilled outputs.
Visible errors in the student's predictions show that our distillation method presented in this paper does not solve the explicit 2D pose estimation problem in challenging sports data.
However, solving this explicit task is not necessarily required to improve results in downstream tasks that depend on pose.

%% file: tables/weak_pose_stats.tex
\begin{table*}[tbp]
    \centering
    \begin{tabularx}{\textwidth}{cccrrcccrrccrr}
        \toprule
        Dataset&
            \multicolumn{4}{c}{\tennis} &&
            \multicolumn{4}{c}{\fx} &&
            \multicolumn{3}{c}{Diving48} \\
        Score
            & \% All & \% Action & Full & \multicolumn{1}{c}{16-shot} &
            & \% All & \% Action & Full & \multicolumn{1}{c}{16-shot} &
            & \% All & Full & \multicolumn{1}{c}{16-shot} \\
        \midrule
        \multicolumn{1}{l}{\VIPESTAR}
            & - & - & 91.8 & 67.0 &
            & - & - & 90.8 & 75.7 &
            & - & 78.6 & 35.0 \\
        $\geq0.1$
            & 99.4 & 99.2 & 93.4 & 67.5 &
            & 99.7 & 99.5 & 93.9 & 82.9 &
            & 89 & 81.1 & 43.9 \\
        $\geq0.3$
            & 99.2 & 99.0 & 93.0 & 67.7 &
            & 96 & 91 & 93.8 & 84.0 &
            & 62 & 85.7 & 51.8 \\
        $\geq0.5$
            & 97.9 & 97.2 & \textbf{93.4} & 69.5 &
            & 90 & 79 & \textbf{94.6} & \textbf{84.9} &
            & 38 & \textbf{88.9} & \textbf{58.8} \\
        $\geq0.7$
            & 89.9 & 83.9 & 93.3 & \textbf{71.1} &
            & 78 & 61 & 93.9 & 83.0 &
            & 17 & 87.7 & 50.6 \\
        $\geq0.9$
            & 1.5 & 1.3 & 91.2 & 65.4 &
            & 17 & 8 & 93.1 & 79.9 &
            & $<$1 & 73.8 & 25.7 \\
        \bottomrule
    \end{tabularx}
    % \vspace{-0.08in}
    \caption{{\bf Top-1 accuracy on action recognition using VI-\OURMETHOD when varying the weak-pose selection threshold.}
             For consistency, all results are using the BiGRU (\autoref{sub:gru}).
             Excluding the least confident poses improves accuracy; these poses are most likely to be incorrect.
             However, setting the threshold too high also decreases accuracy if the supervision becomes too sparse.
             The percent of poses in all frames (\% All) and in action frames (\% Action) that are retained at each threshold is also shown.
             Note: Diving48~\cite{diving48} only contains action frames.
    }
    \label{tab:sparse_pose}
    % \vspace{-1em}
\end{table*}

%% file: tables/nns_vs_gru_action.tex
\begin{table*}[tbp]
    {\small
    \centering
    \begin{tabularx}{\textwidth}{lcccccccc}
        \toprule
        \multicolumn{1}{c}{Training data}
            & \multicolumn{2}{c}{4-shot}
            & \multicolumn{2}{c}{8-shot}
            & \multicolumn{2}{c}{16-shot}
            & \multicolumn{2}{c}{32-shot} \\
        \multicolumn{1}{c}{Features \textbackslash~Model}
            & BiGRU & NNS & BiGRU & NNS
            & BiGRU & NNS & BiGRU & NNS \\
        \midrule
        % \vspace{-0.9em} \\
        \multicolumn{9}{l}{\hspace{7pt}\fs} \\
        \midrule
        Normalized 2D joints
            & 38.5 $\pm$ 3.7 & 50.8 $\pm$ 6.1
            & 60.1 $\pm$ 4.5 & 65.3 $\pm$ 4.5
            & 72.5 $\pm$ 3.9 & 71.7 $\pm$ 3.9
            & 89.7 $\pm$ 0.9 & 79.7 $\pm$ 1.8 \\
        \textbf{(Ours)} 2D-\OURMETHOD
            & 43.2 $\pm$ 5.2 & 50.7 $\pm$ 5.8
            & 66.1 $\pm$ 1.1 & 70.3 $\pm$ 3.7
            & 74.4 $\pm$ 3.0 & 75.7 $\pm$ 1.5
            & 90.8 $\pm$ 1.9 & 84.1 $\pm$ 1.2 \\
        \VIPESTAR
            & 51.1 $\pm$ 3.0 & 64.3 $\pm$ 5.0
            & 69.7 $\pm$ 2.9 & 75.7 $\pm$ 3.6
            & 80.5 $\pm$ 3.5 & 78.3 $\pm$ 2.6
            & 91.3 $\pm$ 1.7 & 84.5 $\pm$ 1.3 \\
        \textbf{(Ours)} VI-\OURMETHOD
            & 54.4 $\pm$ 5.0 & \textbf{65.9 $\pm$ 5.5}
            & 71.4 $\pm$ 1.7 & \textbf{78.4 $\pm$ 2.5}
            & 80.2 $\pm$ 1.9 & \textbf{81.1 $\pm$ 2.5}
            & \textbf{92.2 $\pm$ 1.2} & 86.2 $\pm$ 0.7 \\
        \midrule
        % \vspace{-0.9em} \\
        \multicolumn{9}{l}{\hspace{7pt}\tennis} \\
        \midrule
        Normalized 2D joints
            & 48.0 $\pm$ 1.9 & 54.2 $\pm$ 3.4
            & 58.5 $\pm$ 3.0 & 57.0 $\pm$ 5.5
            & 64.4 $\pm$ 2.6 & 63.0 $\pm$ 2.8
            & 69.7 $\pm$ 2.6 & 64.6 $\pm$ 2.3 \\
        \textbf{(Ours)} 2D-\OURMETHOD
            & 53.0 $\pm$ 3.3 & 57.0 $\pm$ 3.4
            & 62.0 $\pm$ 1.7 & 61.3 $\pm$ 4.8
            & 66.9 $\pm$ 1.7 & 65.0 $\pm$ 2.0
            & 71.5 $\pm$ 2.4 & 67.2 $\pm$ 1.5 \\
        \VIPESTAR
            & 61.4 $\pm$ 4.1 & 62.4 $\pm$ 4.4
            & 65.8 $\pm$ 3.4 & 65.6 $\pm$ 3.5
            & 67.0 $\pm$ 2.8 & 68.8 $\pm$ 4.3
            & 73.2 $\pm$ 2.3 & 70.1 $\pm$ 2.0 \\
        \textbf{(Ours)} VI-\OURMETHOD
            & \textbf{63.9 $\pm$ 6.1} & 62.4 $\pm$ 4.5
            & 65.5 $\pm$ 4.5 & \textbf{66.1 $\pm$ 3.5}
            & \textbf{71.1 $\pm$ 2.4} & 68.4 $\pm$ 3.5
            & \textbf{76.3 $\pm$ 2.0} & 70.3 $\pm$ 1.8 \\
        \midrule
        % \vspace{-0.9em} \\
        \multicolumn{9}{l}{\hspace{7pt}\fx} \\
        \midrule
        Normalized 2D joints
            & 37.6 $\pm$ 1.2 & 38.0 $\pm$ 1.9
            & 54.8 $\pm$ 2.6 & 45.8 $\pm$ 1.2
            & 65.6 $\pm$ 0.9 & 52.8 $\pm$ 1.4
            & 75.3 $\pm$ 0.9 & 59.0 $\pm$ 0.6 \\
        \textbf{(Ours)} 2D-\OURMETHOD
            & 51.2 $\pm$ 1.0 & 47.4 $\pm$ 2.1
            & 70.0 $\pm$ 1.2 & 54.9 $\pm$ 1.5
            & 82.7 $\pm$ 0.6 & 63.9 $\pm$ 1.4
            & 88.8 $\pm$ 0.8 & 69.7 $\pm$ 0.5 \\
        \VIPESTAR
            & 49.7 $\pm$ 0.7 & 43.0 $\pm$ 1.7
            & 62.5 $\pm$ 2.1 & 49.1 $\pm$ 0.9
            & 75.7 $\pm$ 0.4 & 54.3 $\pm$ 1.2
            & 81.8 $\pm$ 0.5 & 59.7 $\pm$ 1.3 \\
        \textbf{(Ours)} VI-\OURMETHOD
            & \textbf{59.3 $\pm$ 1.9} & 51.0 $\pm$ 1.1
            & \textbf{73.0 $\pm$ 0.6} & 57.1 $\pm$ 1.3
            & \textbf{84.9 $\pm$ 0.5} & 65.4 $\pm$ 1.5
            & \textbf{89.1 $\pm$ 0.6} & 70.6 $\pm$ 0.7 \\
        \midrule
        % \vspace{-0.9em} \\
        \multicolumn{9}{l}{\hspace{7pt}Diving48} \\
        \midrule
        Normalized 2D joints
            & 12.6 $\pm$ 1.2 & 13.3 $\pm$ 1.4
            & 13.3 $\pm$ 1.2 & 15.3 $\pm$ 0.8
            & 25.5 $\pm$ 3.5 & -
            & 44.2 $\pm$ 0.9 & - \\
        \textbf{(Ours)} 2D-\OURMETHOD
            & 27.6 $\pm$ 2.6 & 18.4 $\pm$ 2.4
            & 29.4 $\pm$ 1.2 & 22.8 $\pm$ 1.4
            & 57.6 $\pm$ 6.5 & -
            & 76.6 $\pm$ 0.9 & -\\
        \VIPESTAR
            & 17.0 $\pm$ 1.6 & 12.9 $\pm$ 1.6
            & 18.8 $\pm$ 1.0 & 16.1 $\pm$ 1.3
            & 35.0 $\pm$ 4.5 & -
            & 53.2 $\pm$ 1.4 & - \\
        \textbf{(Ours)} VI-\OURMETHOD
            & \textbf{29.2 $\pm$ 2.5} & 16.9 $\pm$ 2.1
            & \textbf{34.0 $\pm$ 1.2} & 21.2 $\pm$ 1.0
            & \textbf{58.8 $\pm$ 3.6} & -
            & \textbf{76.7 $\pm$ 0.8} & - \\
        \bottomrule
    \end{tabularx}
    }
    % \vspace{-0.08in}
    \caption{{\bf Results with NNS, under $L_2$ distance and DTW, compared to the BiGRU in the few-shot setting.}
    NNS can perform competitively in label-poor settings, though the results are dataset dependent.
    We did not evaluate NNS on Diving48 past $k=8$ due to the large number of classes (48), longer average clip length, and the inference time scaling linearly with the number of training examples.
    }
    \label{tab:fs_nns_dtw}
    % \vspace{-1em}
\end{table*}

%% file: tables/detection_threshold.tex
\begin{table*}[p]
    \centering
    \begin{tabularx}{0.74\textwidth}{lrrrrrrrrrr}
        \toprule
        & \multicolumn{9}{c}{Figure skating jumps} \\
        \multicolumn{1}{c}{Activation threshold} &
            \multicolumn{1}{c}{0.1} &
            \multicolumn{1}{c}{0.2} &
            \multicolumn{1}{c}{0.3} &
            \multicolumn{1}{c}{0.4} &
            \multicolumn{1}{c}{0.5} &
            \multicolumn{1}{c}{0.6} &
            \multicolumn{1}{c}{0.7} &
            \multicolumn{1}{c}{0.8} &
            \multicolumn{1}{c}{0.9} \\
        \midrule
        Pretrained R3D~\cite{r3d}
            & 23.3 & \multicolumn{1}{|r|}{23.1} & 18.2 & 16.6 & 14.2
            & 12.8 & 10.0 & 7.4 & 5.9 \\
        Normalized 2D Joints
            & 57.1 & \multicolumn{1}{|r|}{53.4} & 50.4 & 46.9 & 42.2
            & 37.9 & 33.3 & 27.0 & 20.3 \\
        2D-\OURMETHOD
            & \textbf{63.3} & \multicolumn{1}{|r|}{\textbf{61.5}} & 58.7 & 56.3 & 55.2
            & \textbf{53.7} & \textbf{51.8} & 49.2 & 44.4 \\
        \VIPESTAR
            & 61.5 & \multicolumn{1}{|r|}{59.3} & 58.1 & \textbf{57.8} & \textbf{56.1}
            & 52.0 & 50.0 & 44.6 & 40.0 \\
        VI-\OURMETHOD
            & 61.7 & \multicolumn{1}{|r|}{60.7} & \textbf{59.6} & 57.5 & 54.9
            & 53.0 & 51.2 & \textbf{49.9} & \textbf{45.7} \\
        \midrule
        \midrule
        & \multicolumn{9}{c}{Tennis swings at 200 ms} \\
        \multicolumn{1}{c}{Activation threshold} &
            \multicolumn{1}{c}{0.05} &
            \multicolumn{1}{c}{0.1} &
            \multicolumn{1}{c}{0.15} &
            \multicolumn{1}{c}{0.2} &
            \multicolumn{1}{c}{0.25} &
            \multicolumn{1}{c}{0.3} &
            \multicolumn{1}{c}{0.35} &
            \multicolumn{1}{c}{0.4} &
            \multicolumn{1}{c}{0.45} \\
        \midrule
        Pretrained R3D~\cite{r3d}
            & 32.1 & 31.5 & 30.9 & \multicolumn{1}{|r|}{29.9} & 28.5
            & 27.4 & 26.4 & 25.1 & 22.3 \\
        Normalized 2D Joints
            & 46.8 & 45.3 & 44.4 & \multicolumn{1}{|r|}{43.7} & 43.6
            & 43.1 & 41.8 & 40.1 & 37.7 \\
        2D-\OURMETHOD
            & 51.2 & 52.1 & 53.5 & \multicolumn{1}{|r|}{54.0} & 54.7
            & 54.5 & 53.6 & 52.0 & 49.0 \\
        \VIPESTAR
            & 49.6 & 50.1 & 50.8 & \multicolumn{1}{|r|}{51.2} & 51.8
            & 52.1 & 51.5 & 50.3 & 48.0 \\
        VI-\OURMETHOD
            & \textbf{55.6} & \textbf{56.3} & \textbf{57.8} & \multicolumn{1}{|r|}{\textbf{58.6}} & \textbf{59.3}
            & \textbf{59.8} & \textbf{59.7} & \textbf{58.9} & \textbf{56.9} \\
        \bottomrule
    \end{tabularx}
    % \vspace{-0.08in}
    \caption{{\bf Few-shot action detection: Average precision (AP) at $\text{tIoU}=0.5$ with various frame-level activation thresholds.}
        These per-frame activations are produced by the BiGRU ensemble described in~\autoref{sec:impl:fewshotdetection}, and consecutive activations above the threshold are predicted as actions. In~\autoref{tab:detection}, we use a threshold of 0.2.}
    \label{tab:detection_threshold}
    % \vspace{-1em}
\end{table*}

%% file: tables/other_action_arch.tex
\begin{table*}[tp]
    \centering
    \begin{tabularx}{\textwidth}{lccccccccccc}
        \toprule
        \multicolumn{1}{c}{Dataset}
            & \multicolumn{2}{c}{\fs} &
            & \multicolumn{2}{c}{\tennis} &
            & \multicolumn{2}{c}{\fx} &
            & \multicolumn{2}{c}{Diving48} \\
        \multicolumn{1}{c}{Architecture \textbackslash~Features}
            & \VIPESTAR & VI-\OURMETHOD &
            & \VIPESTAR & VI-\OURMETHOD &
            & \VIPESTAR & VI-\OURMETHOD &
            & \VIPESTAR & VI-\OURMETHOD \\
        \midrule
        \hspace{-4pt}NNS (w/ DTW) [\autoref{sub:nns_dtw}]
            & 90.6 & 92.7 &
            & 89.1 & 88.6 &
            & 71.8 & 81.2 &
            & - & - \\
        \hspace{-4pt}CNN~\cite{textcnn}
            & 93.8 & 96.0 &
            & 91.6 & 93.0 &
            & 87.8 & 93.4 &
            & 58.8 & 81.3 \\
        \hspace{-4pt}BiLSTM
            & 97.7 & 98.1 &
            & 92.2 & \textbf{93.4} &
            & 90.9 & 94.3 &
            & 77.9 & 88.2 \\
        \hspace{-4pt}BiLSTM (w/ attn)
            & 97.3 & 97.9 &
            & 90.7 & 92.0 &
            & 88.8 & 93.9 &
            & 76.8 & 87.5\\
        \hspace{-4pt}BiGRU [\autoref{sub:gru}]
            & 96.8 & 97.4 &
            & 91.8 & 93.3 &
            & 90.8 & \textbf{94.6} &
            & 78.6 & \textbf{88.6} \\
        \hspace{-4pt}BiGRU (w/ attn)
            & 96.8 & \textbf{98.3} &
            & 91.1 & 92.5 &
            & 89.5 & 94.3 &
            & 77.5 & 88.0 \\
        \bottomrule
    \end{tabularx}
    % \vspace{-0.08in}
    \caption{{\bf Action recognition architectures: Top-1 accuracy using \VIPESTAR and VI-\OURMETHOD features in the full supervision setting.}
    We experimented with a number of standard architectures for classifying sequences of pose features.
    The CNN is based on early work on text classification with word vectors~\cite{textcnn}.
    The BiLSTM is similar to the BiGRU described in~\autoref{sub:gru}.
    For the BiLSTM and BiGRU with attention, we use an attention mechanism similar to~\cite{pytorchseq2seq}.
    Results are often similar when comparing across architectures, showing that the  improvement from VI-\OURMETHOD is not reliant on the downstream architecture.
    For consistency, we use the BiGRU, without attention, for the main results in the paper.}
    \label{tab:other_action_arch}
    % \vspace{-1em}
\end{table*}

%% file: tables/gsm_crop_diving48.tex
\begin{table}[tp]
    \centering
    \begin{tabularx}{0.81\columnwidth}{crrr}
        \toprule
        \multicolumn{1}{c}{$k$} & \multicolumn{1}{c}{Not cropped} & \multicolumn{1}{c}{Cropped} & \multicolumn{1}{c}{Difference} \\
        \midrule
        8 & 9.5 $\pm$ 1.3 & 15.1 $\pm$ 0.8 & +5.5 \\
        16 & 21.8 $\pm$ 1.5  & 33.0 $\pm$ 0.9 & +11.2 \\
        32 & 49.5 $\pm$ 3.1 & 59.3 $\pm$ 2.4 & +9.8 \\
        64 & 72.6 $\pm$ 1.0 & 75.6 $\pm$ 1.2 & +3.0 \\
        \bottomrule
    \end{tabularx}
    % \vspace{-0.08in}
    \caption{{\bf GSM~\cite{gsm} on Diving48~\cite{diving48}, with and without cropping, in the $k=8$ to $k=64$ shot settings.} GSM with subject cropping is the stronger baseline and is used in the few-shot action recognition experiments in~\autoref{sub:few_shot_action}.}
    \label{tab:gsm_few_shot}
    % \vspace{-0.5em}
\end{table}

%% file: supplemental/dataset.tex
\input{tables/dataset_stats}

\section{Additional Dataset Details}
\label{sec:dataset_details}

This section provides additional details about the fine-grained sports video datasets used in the results section.

\heading{Figure skating} is a new dataset that contains the jumps in 371 singles short programs.
Because professional skaters often repeat the same routine in a competitive season, all performances from 2018 are held out for testing.

The six jump types that occur in the \fs dataset are:
Axel, flip, loop, Lutz, Salchow, and toe-loop (see~\autoref{tab:fs_dist}).
The labels are verified against the ISU's~\cite{isu} publicly accessible scoring data.
For the classification task,
the average label duration is 3.3 seconds and includes the poses from
before taking off and after landing.

\heading{Tennis} consists of Vid2Player's~\cite{vid2player} swing annotations in nine matches.
For action recognition, \tennis has seven swing types: forehand topspin, backhand topspin, forehand slice, backhand slice, forehand volley, backhand volley, and overhead.
Note that the distribution of actions in tennis is unbalanced, with forehand topspin being the most common.
Serves are intentionally excluded from the action recognition task because they always occur at the start of points and do not need to be classified.
For swing detection, however, serves are included.

All action recognition models receive a one second interval, centered around the frame of ball contact for the swing.

\heading{Floor exercise.} We use the videos, labels, and official train/validation split from the floor exercise event of FineGym99~\cite{finegym}.
We focus on floor exercises (\fx) because the data is readily tracked and because the~\cite{finegym} authors report accuracies on this subset.
Because actions are often short, for each action, we extracted frames from 250 ms prior to the annotated start time to the end time, and we use these frames as the inputs to our methods and the baselines.

\heading{Diving48~\cite{diving48}} contains both individual and synchronized diving.
We use the standard train/validation split.
For synchronized diving, we track either diver as the subject and tracks can flicker between divers due to missed detections.
Tracking is the most challenging in this dataset because of the low resolution, motion blur, and occlusion upon entering the water.
Also, because the clips are short, it is more difficult to initialize tracking heuristics that utilize periods of video before and after an action, where the athlete is more static and can be more easily detected and identified.

\subsection*{Subject Tracking}
To focus on the athletes, we introduce subject tracking to the figure skating, floor exercises~\cite{finegym}, and Diving48~\cite{diving48} datasets.
Our annotations are created with off-the-shelf person detection and tracking algorithms.
First, we run a Mask R-CNN detector with a ResNeXt-152-32x8d backbone~\cite{detectron2} on
every frame to detect instances of people.
We use heuristics such as ``the largest person in the frame'' (e.g., in figure skating, floor exercise, and diving)
and ``upside down pose'' (e.g., in floor exercise and diving) to select the athlete.
These selections are tracked across nearby frames with bounding box
intersection-over-union, SORT~\cite{sort}, and OpenCV~\cite{opencv} object tracking (CSRT~\cite{csrt}) when detections are missed.
This heuristic approach is similar to the one taken by the authors of Vid2Player~\cite{vid2player}.

Example images of tracked and cropped athletes are shown in~\autoref{fig:example_crops}.
We run pose estimation on the pixels contained in and around the tracked boxes.

%% file: tables/dataset_stats.tex
\begin{table}[t]
    \centering
    \begin{tabularx}{0.36\columnwidth}{lr}
        \toprule
        \multicolumn{1}{c}{Class} & Count \\
        \midrule
        Axel & 371 \\
        Flip & 179 \\
        Loop & 94 \\
        Lutz & 244 \\
        Salchow & 61 \\
        Toe-loop & 497 \\
        \midrule
        Total & 1,446 \\
        \bottomrule
    \end{tabularx}
    % \vspace{-0.08in}
    \caption{{\bf Distribution of action classes in \fs.}}
    \label{tab:fs_dist}
    \vspace{2em}
% \end{table}
% \begin{table}[t]
    \centering
    \begin{tabularx}{0.505\columnwidth}{lr}
        \toprule
        \multicolumn{1}{c}{Class} & Count \\
        \midrule
        Backhand slice & 812 \\
        Backhand topspin & 3,134 \\
        Backhand volley & 140 \\
        Forehand slice & 215 \\
        Forehand topspin & 3,732 \\
        Forehand volley & 123 \\
        Overhead & 87 \\
        \midrule
        Total & 8,243 \\
        \bottomrule
    \end{tabularx}
    % \vspace{-0.08in}
    \caption{{\bf Distribution of action classes in \tennis.}}
    \label{tab:tennis_dist}
    % \vspace{-1em}
\end{table}

%% file: supplemental/vipestar.tex
\section{\VIPESTAR Details}
\label{sec:jamesvipe}

We provide details of~\VIPESTAR, which is used as the teacher for our view-invariant  VI-\OURMETHOD student.
\VIPESTAR is used because the evaluation code and documentation for Pr-VIPE~\cite{prvipe}
is not released at the time of development.
The experiments in this section are to demonstrate that
\VIPESTAR is a suitable substitute, based
on~\cite{prvipe}'s evaluation on coarse-grained action recognition.

\heading{Overview.}
View-invariant pose embedding (VIPE) methods embed 2D joints such that different camera views of the same pose
in 3D are similar in the embedding space. \VIPESTAR is trained via 3D lifting
to canonicalized features (w.r.t. rotation and body shape).
We designed \VIPESTAR to train on multiple (publicly available) datasets with differing 3D
joint semantics; we use Human3.6M~\cite{human36m} as well as synthetic pose data
from 3DPeople~\cite{3dpeople}, AMASS~\cite{amass}, and NBA2K~\cite{nba2k}.

\heading{Inputs.}
\VIPESTAR learns view-invariant embeddings by regressing 3D joint
features from 2D joint pose.
The 2D joint inputs are the 13 COCO~\cite{coco}
keypoints (excluding eyes and ears) normalized as in~\cite{prvipe}.
To obtain canonicalized 3D features, first, we rotate the 3D pose around the vertical-axis,
aligning the torso-normal vector to the depth-axis.
Then, we normalize each
joint as two unit length offsets from its parent and from the hip (centered to 0).
We also concatenate the cosine bone angle at each 3D joint.
These transformations standardize 3D poses with respect to body appearance and camera view.

\heading{Model.} \VIPESTAR uses a similar neural network backbone to~\cite{3dliftingbaseline,prvipe} and is trained with two losses:
\begin{itemize}
    \item {\em 3D feature reconstruction loss.}
    We use a fully-connected decoder that takes embeddings as input.
    This decoder is discarded after training. To support multi-task training
    with 3D datasets with different ground-truth joint semantics, we specialize
    the output layer weights for each dataset.

    \item {\em Contrastive embedding loss.}
    We minimize the pairwise $L_2$ distance between embeddings of different 2D views of the same 3D
    pose (positive pairs).
    We also negatively sample pairs of 2D poses, corresponding to different
    3D poses in each action sequence, and maximize their embedding distance.
    Two 3D poses are considered to be different if one of their
    joint-bone angles differs by 45$^\circ$ or more.
\end{itemize}

\heading{Substitute for Pr-VIPE.}
We compare \VIPESTAR's performance to the coarse-grained action recognition results reported by~\cite{prvipe,cvmim}
on the Penn Action~\cite{pennaction} dataset. Our results suggest parity with Pr-VIPE when trained with Human3.6M only
and a small improvement from extra synthetic data.
\VIPESTAR has 98.2\% top-1 accuracy (compared to 98.4\%, the best result for Pr-VIPE~\cite{cvmim})
when trained on the same subjects of the
Human3.6M dataset and using nearest-neighbor search as the action recognition method (see~\autoref{sub:nns_dtw}).
\VIPESTAR obtains 98.6\% accuracy when trained with extra synthetic 3D data.
The saturated accuracies of \VIPESTAR, Pr-VIPE~\cite{prvipe}, and other prior
work~\cite{cvmim} on the Penn Action dataset suggest that more challenging datasets,
such as fine-grained sports, are needed to evaluate new techniques.

For fine-grained action recognition in sports, additional synthetic 3D data improves
\VIPESTAR (\autoref{tab:more_data_for_vipe}).
This is especially notable on \fx and Diving48, which contain a variety
of poses that are not well represented by Human3.6M.
We use \VIPESTAR, improved with the synthetic 3D data, as the teacher for all of
our VI-\OURMETHOD experiments.

\input{tables/vipe_synthetic_data}

% Big figures
\input{figures/explicit_2d_pose}
\input{figures/example_crops}

%% file: tables/vipe_synthetic_data.tex
\begin{table}[t]
    \centering
    \begin{tabularx}{0.825\columnwidth}{lccr}
        \toprule
        & \multicolumn{3}{c}{\VIPESTAR training data} \\
        \multicolumn{1}{c}{Dataset} & \multicolumn{1}{c}{Human3.6M} & \multicolumn{1}{c}{All} & \multicolumn{1}{c}{Difference} \\
        \midrule
        \fs & 95.8 & 96.8 & +1.0 \\
        \tennis & 91.9 & 91.8 & -0.1 \\
        \fx & 87.7 & 90.8 & +3.1\\
        Diving48 & 66.8 & 76.8 & +10.0\\
        \bottomrule
    \end{tabularx}
    % \vspace{-0.08in}
    \caption{{\bf Effect of additional synthetic 3D data (all) for \VIPESTAR on fine-grained sports datasets.} Top-1 accuracy on fine-grained sports action recognition is shown. The improvement is largest on \fx and Diving48, which differ the most from the common poses in Human 3.6M~\cite{human36m}. We use \VIPESTAR trained with all of the 3D data as the teacher for VI-\OURMETHOD and the \VIPESTAR baselines in~\autoref{sec:action_recognition}. Note that vertically augmenting \VIPESTAR for Diving48, as described in~\autoref{sec:action_recognition}, further increases accuracy to 78.6\% (over the 76.8\% shown above).
    }
    \label{tab:more_data_for_vipe}
    % \vspace{-1em}
\end{table}

%% file: figures/explicit_2d_pose.tex
\begin{figure*}[p]
    \centering
    \includegraphics[width=\textwidth]{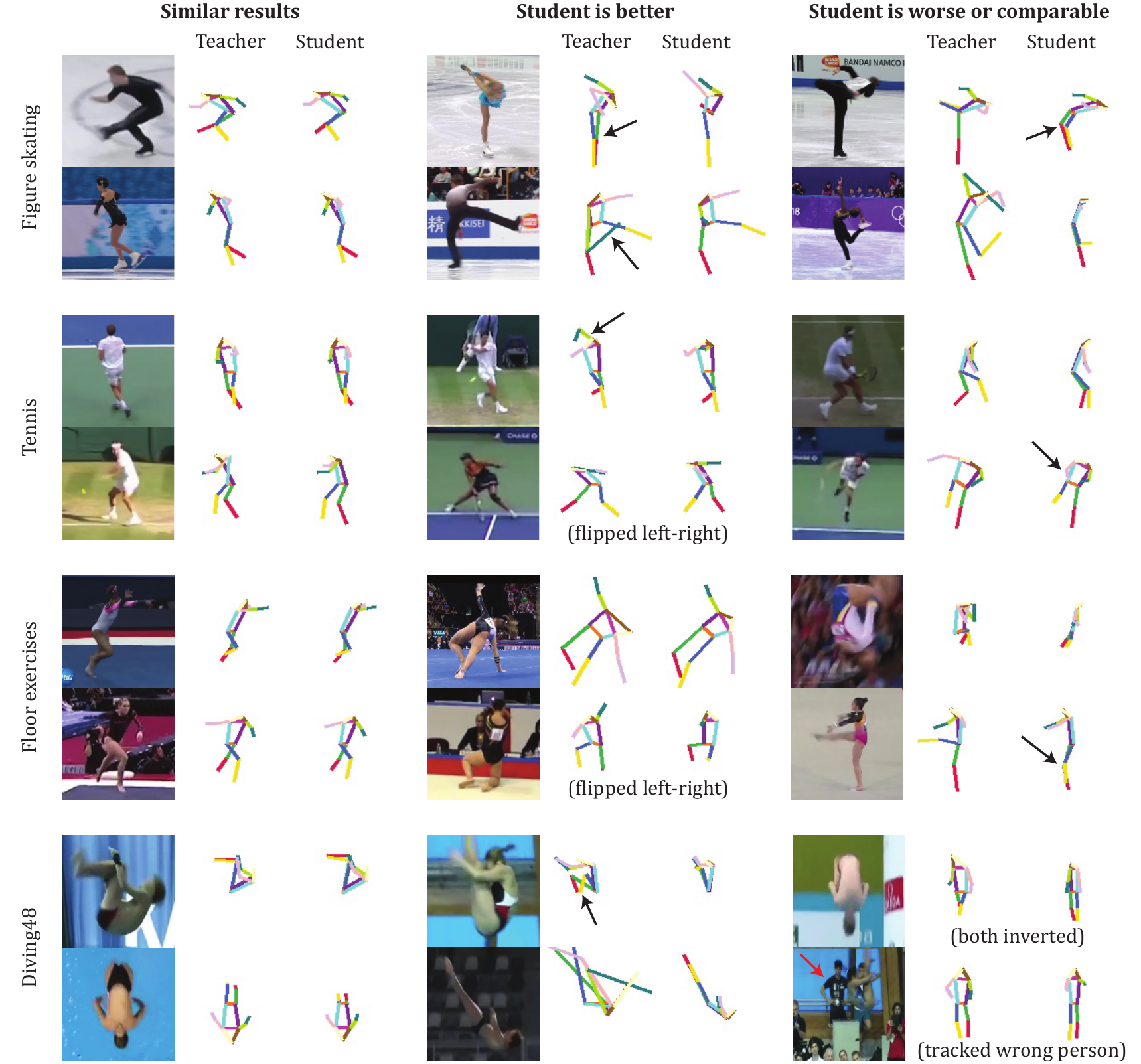}
    \vspace{-0.2in}
    \caption{
        {\bf Qualitative examples of distilled 2D joints.}
        To investigate whether distillation improves the more general, explicit 2D pose estimation problem,
        we visualize the output of an ablated student trained to distill 2D poses, without the motion component. 
        The sole learning objective in this experiment is to mimic the normalized 2D joint offsets produced by the teacher.
        As seen above, the student's output often is similar to the teacher's, especially when the teacher performs well (left column).
        The student can sometimes produce more plausible results by avoiding extreme errors by the teacher, such as inverting left and right or when pose keypoints are jumbled (middle column).
        However, as the example failures in the right column show, this student is far from perfect.
        Ground-truth 2D pose is not available for these datasets, making further quantitative evaluation difficult.
    }
    \label{fig:pose_examples}
    % \vspace{-1em}
\end{figure*}

%% file: figures/example_crops.tex
\begin{figure*}[p]
    \centering
    \begin{subfigure}{\textwidth}
        \includegraphics[width=\textwidth]{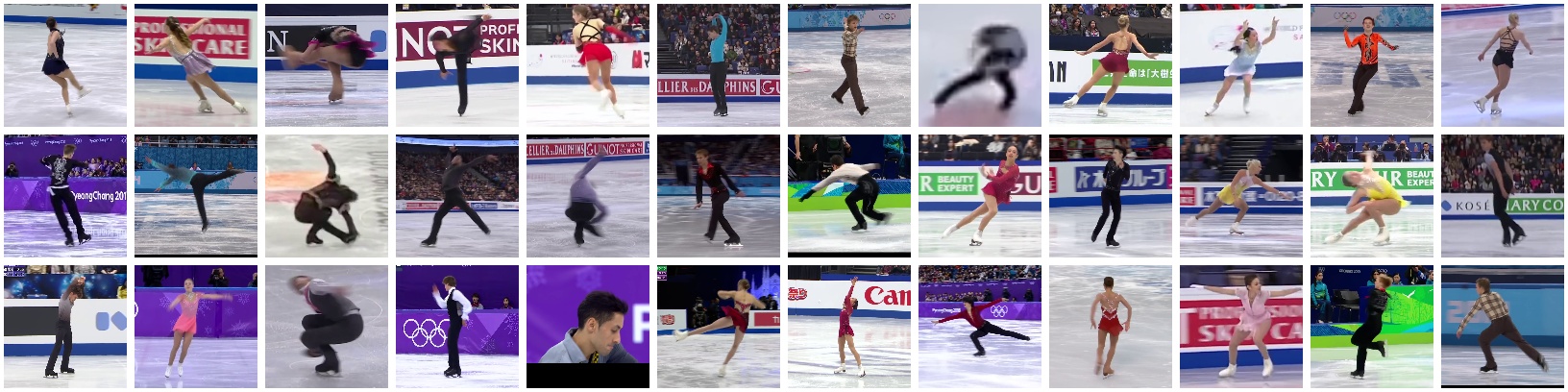}
        \caption{Figure skating. There is large variation in camera view and clothing appearance.}
        \vspace{1em}
    \end{subfigure}
    \begin{subfigure}{\textwidth}
        \includegraphics[width=\textwidth]{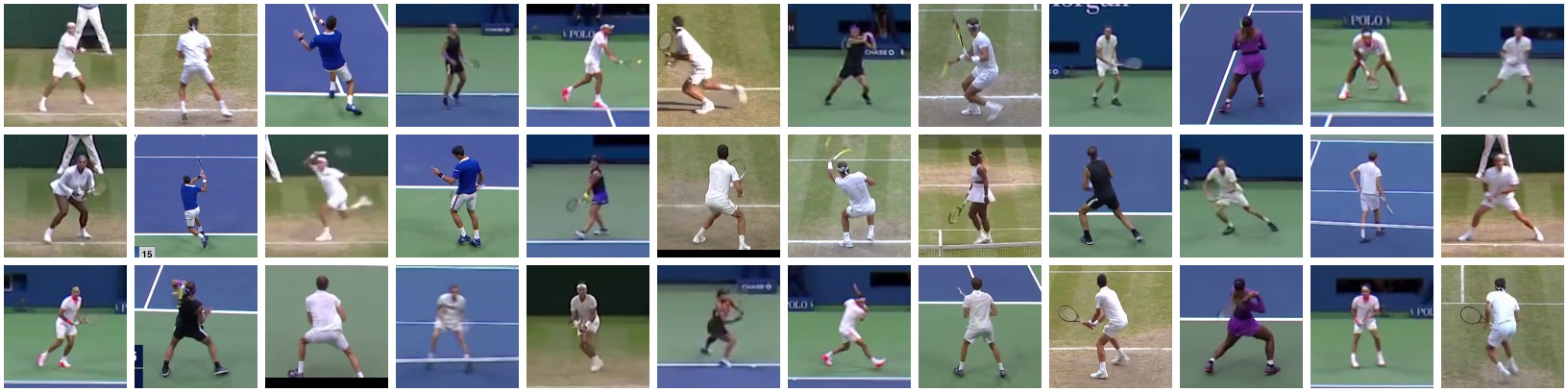}
        \caption{Tennis. Note the bimodality in camera angle (foreground and background) and court (US Open hard court vs. Wimbledon grass).}
        \vspace{1em}
    \end{subfigure}
    \begin{subfigure}{\textwidth}
        \includegraphics[width=\textwidth]{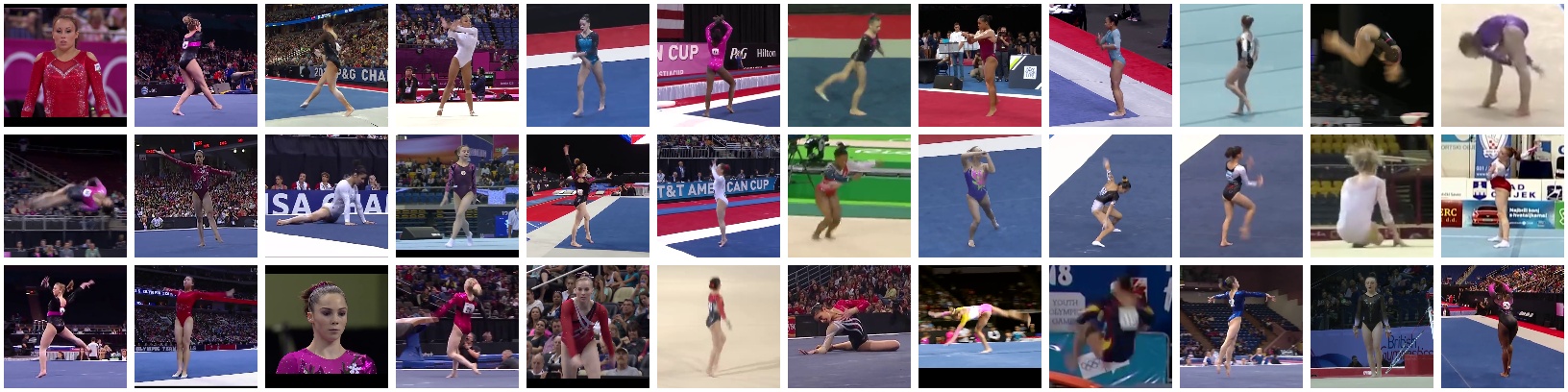}
        \caption{Floor exercise. There is large variation in camera view, background, and clothing colors.}
        \vspace{1em}
    \end{subfigure}
    \begin{subfigure}{\textwidth}
        \includegraphics[width=\textwidth]{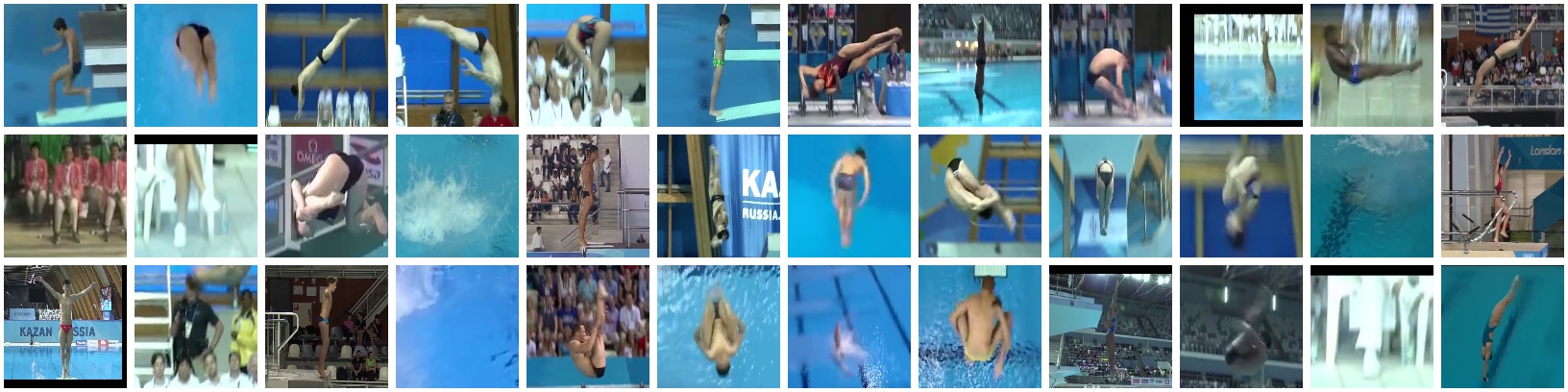}
        \caption{Diving48. There are frequent errors and lapses in pose tracking, especially after entry into the water.}
    \end{subfigure}
    \caption{
        {\bf Examples of cropped athletes}, based on tracking in the four sports datasets. 
    }
    \label{fig:example_crops}
\end{figure*}